\pdfoutput=1

\documentclass[11pt]{article}
\usepackage{own}

\usepackage{times}
\usepackage{latexsym}

\usepackage[T1]{fontenc}

\usepackage[utf8]{inputenc}

\usepackage{microtype}

\usepackage{inconsolata}
\usepackage{caption}

\usepackage{graphicx} 
\usepackage{amsmath,amsthm,amssymb}
\usepackage{mathtools}
\usepackage{booktabs}
\usepackage{float} 
\usepackage{caption}
\usepackage{subcaption}
\usepackage{tabularray}

\newcommand{\x}{\boldsymbol{x}}

\definecolor{lightgray}{HTML}{acacac}

\usepackage{todonotes}
\usepackage{comment}
\usepackage{tabularx}
\usepackage{adjustbox}

\usepackage{multicol}

\usepackage{multirow}

\usepackage{setspace}
\usepackage{enumitem}
\setlist{noitemsep,topsep=0pt,parsep=0pt,partopsep=0pt,leftmargin=8pt}
\newcolumntype{C}[1]{>{\centering\let\newline\\\arraybackslash\hspace{0pt}}m{#1}}

\newcommand*{\myfont}{\fontfamily{phv}\selectfont}
\DeclareTextFontCommand{\textmyfont}{\myfont}

\title{Explaining Text Similarity in Transformer Models}

\author{Alexandros Vasileiou$^{1}$ \quad  Oliver Eberle$^{1,2}$   \\\\
  $^{1}$Machine Learning Group, Technische Universität Berlin, Berlin, Germany \\
  $^{2}$BIFOLD -- Berlin Institute for the Foundations of Learning and Data, Berlin, Germany \\
\texttt{\small vasileiou.a@gmx.de \quad oliver.eberle@tu-berlin.de}
}
\begin{document}
\definecolor{lightgray}{HTML}{acacac}

\maketitle
\begin{abstract}
As Transformers have become state-of-the-art models for natural language processing (NLP) tasks, the need to understand and explain their predictions is increasingly apparent. Especially in unsupervised applications, such as information retrieval tasks, similarity models built on top of foundation model representations have been widely applied. However, their inner prediction mechanisms have mostly remained opaque. Recent advances in explainable AI have made it possible to mitigate these limitations by leveraging improved explanations for Transformers through layer-wise relevance propagation (LRP). Using BiLRP, an extension developed for computing second-order explanations in bilinear similarity models, we investigate which feature interactions drive similarity in NLP models. We validate the resulting explanations and demonstrate their utility in three corpus-level use cases, analyzing grammatical interactions, multilingual semantics, and biomedical text retrieval.
Our findings contribute to a deeper understanding of different semantic similarity tasks and models, highlighting how novel explainable AI methods enable in-depth analyses and corpus-level insights.

\end{abstract}

\section{Introduction}
Modern foundation models provide flexible text representations that enable the detection of semantic structure in vast amounts of unlabeled data. While traditionally supervised settings have taken a predominant role for NLP research, many widely used unsupervised tasks have received growing attention. 
The similarity structure of text embeddings herein provides a central starting point for many tasks, including information retrieval, semantic search, ranking and knowledge extraction \cite{ir:frakes-data-structures, ie:10.1145/1089815.1089817, ir:jansen-ir-constructs}, clustering \cite{ aggarwal2012survey}, and visualization \cite{JMLR:v9:vandermaaten08a, JMLR:v11:venna10a, McInnes2018}.
In the context of language generation, the retrieval-augmented generation (RAG) \cite{rag_lewis2020} approach computes semantic closeness to index relevant text data, resulting in numerous new information retrieval systems \cite{gao2023retrieval}.
Furthermore, as the number of models and tasks increases, the need for quantitative evaluation of embeddings has become apparent, leading to the introduction of embedding benchmarks \cite{thakur2021beir, muennighoff2022mteb}.

Complementary to the evaluation of nominal performance, the field of explainable AI aims to provide insights about the inner model mechanisms by highlighting relevant features for a specific prediction \cite{MONTAVON20181, danilevsky-etal-2020-survey, Vilone2021, XAIreview2021}. 
Explanations play a crucial role in verifying that predictions are grounded in task-relevant features, fostering trust and verifiability into complex machine learning models, and enabling the discovery of data patterns and novel insights. 
In this context, the prediction of supervised models can often be explained using heatmaps over input features. Beyond heatmaps, specific explanation approaches have been proposed in the context of unsupervised models \cite{Montavon2022, kaufmann2022} and higher-order explanations \cite{Eberle2020, schnake2020higher, fumagalli2023shapiq}. In this paper, we focus on Transformer-based similarity models that motivate the use of second-order attributions to highlight feature interactions. Our main contributions are as follows: 

\begin{itemize}
    \item We analyze Transformer-based similarity models within the framework of second-order explanations using BiLRP, highlighting the interaction between tokens.
    \item We evaluate explanations through a purposely designed similarity task for which ground truth interactions are available, and via input perturbations on real-world semantic similarity data.   
    \item We investigate the interaction of relevant tokens across three use cases, revealing  the compositional structure that drives high/low  similarity, thus providing fine-grained insights into sentence representations that are the fundamental concept in many NLP applications but have not been analyzed in the context of  explainable AI.     
    \item  We perform corpus-level analyses,  identifying the parts of speech that models prioritize and illustrating how simple token-matching strategies can lead to inaccurate predictions.     
\end{itemize}
Our implementation is publicly available.\footnote{
\url{https://github.com/alevas/xai_similarity_transformers}}

\section{Related Work}

\paragraph*{Semantic Textual Similarity}
The task of identifying the degree of semantic equivalence between texts is referred to as semantic textual similarity  (STS) \cite{lee2005empirical, semeval12, Chandrasekaran2021}. While two words may be semantically related, e.g. ‘coffee’ and ‘mug’, the STS task focuses on the semantic closeness between text, and in this sense ‘tea’ is considered more similar to ‘coffee’ than ‘mug’ \cite{Chandrasekaran2021}.

\paragraph*{Similarity Models for NLP}
Similarity models are designed to capture the meaning and context of the input texts and provide a quantitative measure of their semantic similarity or relatedness. Commonly used approaches include end-to-end-trained Siamese networks \cite{NIPS1993_288cc0ff, 10.5555/2969033.2969055, neculoiu-etal-2016-learning} and universal encoder models with a pooling layer to extract text summary embeddings \cite{cer-etal-2018-universal}. Combined with Transformers, these approaches have become powerful frameworks to build similarity models. Sentence Transformers, specifically Sentence-BERT (SBERT), \cite{reimers-2019-sentence-bert} have emerged as a flexible and widely-used method to compute  compact text representations. Similarity models typically process  pairs of inputs to compute a similarity score, often based on dot products such as cosine similarity, while extensions to dataset-level dot products have enhanced semantic sentence matching \cite{zhong-etal-2020-extractive} and guided document retrieval in RAG systems \cite{rag_lewis2020}.
Recent efforts in evaluating similarity models and text representations include benchmarks on information retrieval \cite{thakur2021beir} and comparing performance across diverse embedding tasks \cite{muennighoff2022mteb}.

\paragraph*{Explaining Transformers}
The limitations of raw attention scores as explanations have emphasized the need for improved explanation methods for Transformers \cite{jain-wallace-2019-attention, serrano-smith-2019-attention}. Notably, aggregating relevant information across attention heads has proven to be a promising direction \cite{abnar-zuidema-2020-quantifying, Chefer_2021_ICCV}, with further empirical evidence supporting the benefits of gradient information \cite{wallace-etal-2019-allennlp, atanasova-etal-2020-diagnostic, chefer2021transformer}. The naive computation of explanatory gradients in Transformers could be further improved by considering the conservation of relevance during backpropagation, which results in a modified layer-wise relevance propagation (LRP) scheme for Transformers \cite{transformerlrp2022}. 

\paragraph*{Interpretable Feature Interaction}
Several works have focused on determining the effect of joint features in supervised classification scenarios. To identify such pair-wise attributions,  multivariate statistics \cite{lasso_interaction_2013, Caruana2015IntelligibleMF}, Hessian-based approaches \cite{DBLP:journals/corr/Janizek2020}, as well as methods inspired by co-operative game theory \cite{tsang2020, shapleyinteraction_2020, fumagalli2023shapiq}, have been proposed. Beyond classification, the interaction between features provides an appropriate level of complexity to assess why a pair of inputs produces a high or low similarity score. To compute joint feature relevance, Hessian$\,\times\,$Product and BiLRP have been proposed \cite{Eberle2020}.  These methods can be derived directly from a deep Taylor decomposition \cite{montavon-pr17} of similarity models and can be seen as extensions of the widely used Gradient$\,\times\,$Input and LRP explanation methods to bilinear models.

\section{Explainable AI for Similarity Models}
In the following section, we briefly outline how the specific structure of deep similarity models motivates the consideration of second-order terms, as well as the need for tailored propagation rules to compute robust and accurate explanations for Transformers.

\subsection{Explainable AI for Similarity Models} Starting from a Taylor expansion of the similarity score $y(\x,\x') = \langle \phi(\x), \phi(\x') \rangle$  with a pair of inputs ($\x$ ,$\x'$), a feature map $\phi: \mathbb{R}^{s\times d} \xrightarrow{} \mathbb{R}^{h}$ and $y$ the predicted dot product similarity, the following description of relevance $R_{ii'}$ assigned to a pair of features $(i,i')$ can be derived \cite{Eberle2020}:
$$
R_{ii'} = x_i x'_{i'}\, [\nabla^2 y(\x,\x')]_{ii'},
$$
where $\nabla^2$ denotes the Hessian containing second-order partial derivatives.

For deep neural networks, these derivatives have been found to be noisy \cite{DBLP:conf/icml/BalduzziFLLMM17, MONTAVON20181}, which motivates the formulation of robustified propagation rules within the LRP framework, resulting in the BiLRP method, which can be computed in the following factored form:
\vspace{-2mm}
\begin{align*}
\text{\small BiLRP}(y,\x,\x') &= \sum_{m=1}^h \text{\small LRP}([\phi_{L} \circ \dots \circ \phi_1]_m,\x)\\[-2mm]
&\quad \quad \otimes \text{\small LRP}([\phi_{L} \circ \dots \circ \phi_1]_m,\x'),
\end{align*}
where $\phi_{L}$ is an intermediate feature map at layer $L$ and $\otimes$ refers to the tensor product between LRP relevance matrices. 
The LRP attribution for neuron $j$ in layer $l$ is computed by pooling over all incoming messages from neurons $k$ in the higher layer $l+1$ :
\vspace{-2mm}
\begin{align*}
R_j^{(l)} &= \sum_k \frac{q_{jk}}{\sum_j q_{jk}}\cdot R_k^{(l+1)}, 
\end{align*}
\noindent with contributions $q_{jk}$ of neuron $j$ to relevance $R_k^{(l+1)}$.
Depending on the type of network layer, different propagation rules have been proposed to compute $q_{jk}$, typically selected to be proportional to the observed neuron activations \cite{lrpoverview}. In Section \ref{sec:propagation_rules}, we introduce specific propagation rules for Transformer models.

The resulting BiLRP explanations assign relevance scores $R_{ii'}$  to each interaction between input features $(x_i, x'_i)$, highlighting in a detailed manner how features interact to produce the similarity prediction. To compute the interactions $R_{ii'}$, one backpropagation pass for each embedding dimension $h$ is required, resulting in $2\times h$ computations for a pair of sentences that can be computed efficiently using automatic differentiation software. For BiLRP, this results in computing multiple LRP explanations and, thus, its robustness is  directly related to the reliable propagation of relevance.

\subsection{Explainable AI for Transformers} \label{sec:propagation_rules}
To compute better explanations for Transformers, leveraging gradient information has proven to be effective. Yet, the non-linear structure of Transformers motivates specific gradient propagation rules to reflect the model prediction more reliably, resulting in more faithful explanations \cite{transformerlrp2022}. The application of these rules does not affect the model's forward predictions but only modifies the gradient computations in the backward pass.

\paragraph{Propagation rules}
The bilinear structure of the query-key-value (QKV) self-attention layers and the layer normalization computations, lead to a break in relevance conservation, which can be addressed using specific propagation rules.
For the \textbf{attention head}, the forward pass can be formulated as $y_j = \sum_i h_i [p_{ij}]_\texttt{.detach()}$, viewing the attention scores $p_{ij}$ 
as a weighting matrix for the current residual stream representation $h_i$ and detaching the associated variable $p_{ij}$ from the computation graph \cite{transformerlrp2022}.
To preserve relevance in \textbf{layer normalization}, the denominator is regarded as a normalization constant resulting in  \mbox{$y_i = (h_i - \mathbb{E}[h])/\big[\sqrt{\epsilon + \text{Var}[h]} \big]_\texttt{.detach()}$}, with expectation $\mathbb{E}$, variance $\text{Var}$ and stabilization parameter $\epsilon$ \cite{transformerlrp2022}.
In addition, specific non-linear activation functions like GeLU break conservation of relevance, which can be addressed by attributing relevance in proportion to the computed activations \cite{structuredxai2022}. 
For all other layers, the LRP-0 propagation rule is applied, redistributing relevance in proportion to the contribution of each input neuron to the computed activation \cite{lrpoverview}.
These resulting propagation rules also match related approaches to explain bilinear LSTM gating \cite{arras-etal-2017-explaining}, and to linearize Transformers \cite{elhage2021mathematical}.

\section{Experiments}
We now introduce the semantic similarity datasets and models used in this paper before proceeding with the evaluation of second-order explanations.

\subsection{Data}

\noindent\textbf{STSb} (Semantic Textual Similarity benchmark)
consists of English text in the form of sentence pairs, extracted  from image captions, news headlines and user forums \cite{datasets:stsb}. A ground truth similarity score was assigned to each text pair as the result of a  human annotation process. In addition, the multilingual STSb (mSTSb) dataset \cite{huggingface:dataset:stsb_multi_mt} provides machine translated sentence pairs in ten languages. \textbf{SICK} (Sentences Involving Compositional Knowledge)  contains sentence pairs with human-annotated similarity scores \cite{marelli-etal-2014-sick}, aiming to assess the performance of models in understanding sentence meaning, compositionality, and related tasks such as textual entailment. \textbf{BIOSSES} consists of a total of one hundred sentence pairs  selected from the `TAC2 Biomedical Summarization Track Training Data Set' \cite{biossess2017}. Each sentence pair was assigned a similarity score based on annotations of five domain experts for the evaluation of biomedical text comparison tasks.

\subsection{Similarity Models}
In our experiments, we consider four Transformer-based architectures and three pooling strategies described in the following.
\paragraph{Transformers}
The widely-used \textbf{BERT} model, initially trained on English text, demonstrates strong performance across various natural language processing tasks \cite{devlin2019bert}. Additionally, it has been extended to 104 languages, resulting in \textbf{mBERT} for cross-lingual applications. Specifically trained on the task of semantic similarity, the \textbf{SBERT} framework provides models that enable the efficient calculation of semantic relatedness, useful for tasks like information retrieval and knowledge extraction \cite{reimers-2019-sentence-bert}. The \textbf{SGPT} \cite{muennighoff2022sgpt} model is built on top of a finetuned GPT-Neo model \cite{gpt-neo}, which is an open source variant of the popular GPT-family and similar to GPT-3. Additional model details are provided in Appendix \ref{app:model_details}.

\paragraph{Pooling}
To summarize token embeddings into fixed size representations, we use the following pooling strategies:
\textbf{CLS-Pooling} uses the CLS token as a fixed-size representation for tasks like text classification in many Transformer models. \textbf{Mean-Pooling} averages over token embeddings to obtain one single representation vector, capturing the overall information of the input sequence. \textbf{QKV-Pooling} uses the QKV-mechanism to compute weighting coefficients before aggregating the token embeddings.

\subsection{Evaluation of Explanations} \label{sec:evaluation}

\begin{figure}[ht!]
    \scriptsize
    \centering
    \begin{tblr}{
        colspec = {XXX},
        hline{2}   = {fg=lightgray, 0.2pt},
        hline{4}   = {fg=lightgray, 0.2pt},
        hline{6}   = {fg=lightgray, 0.2pt},
        hline{8}   = {fg=lightgray, 0.2pt},
        rowsep=2pt, 
        colsep=1.5pt, 
        leftsep=0.5pt, 
        rightsep=0.5pt
      }
    \SetCell[c=1]{l} Ground Truth & &  \hfill ACS\\
     \includegraphics[width=\linewidth]{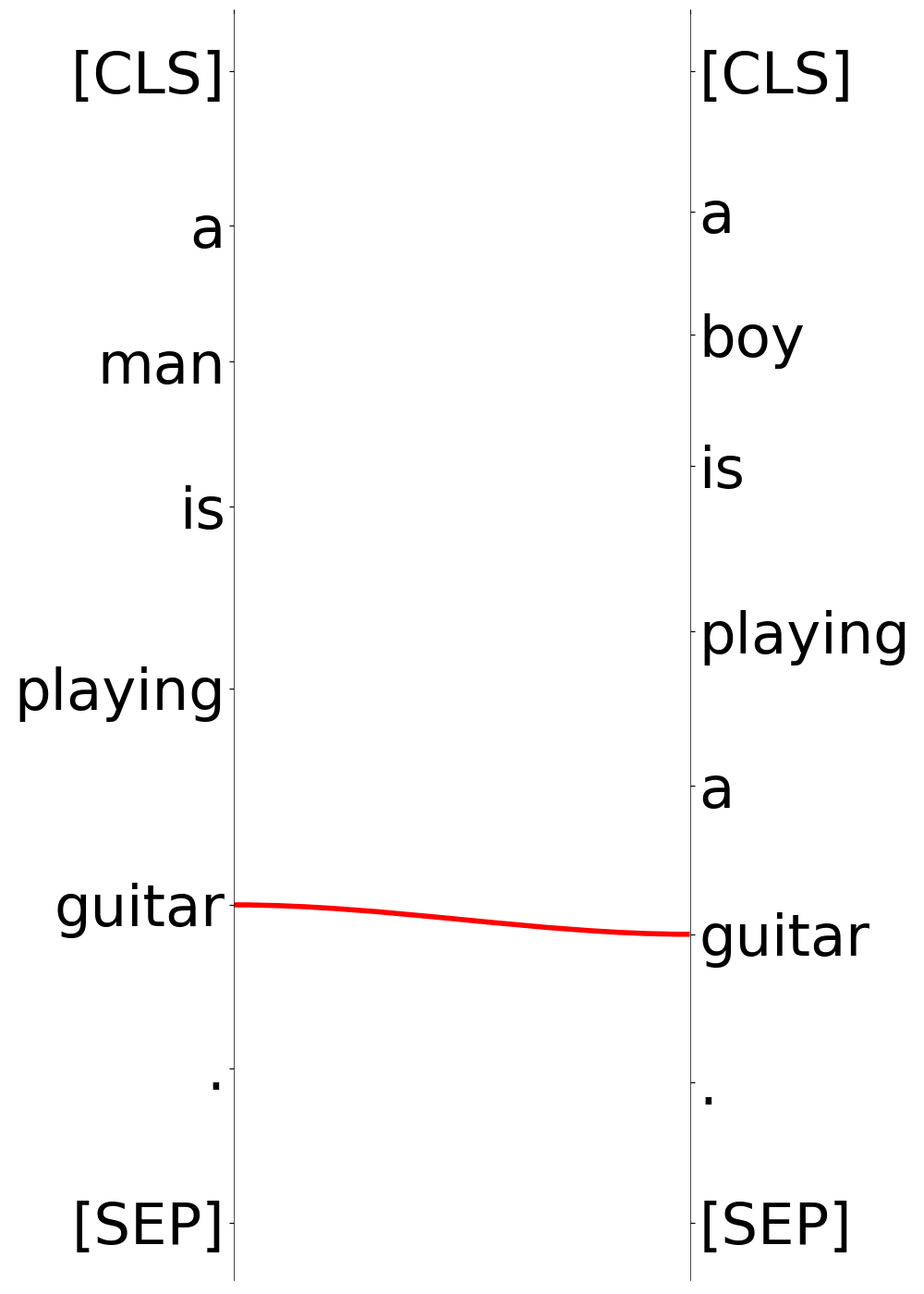} & 
     \includegraphics[width=\linewidth]{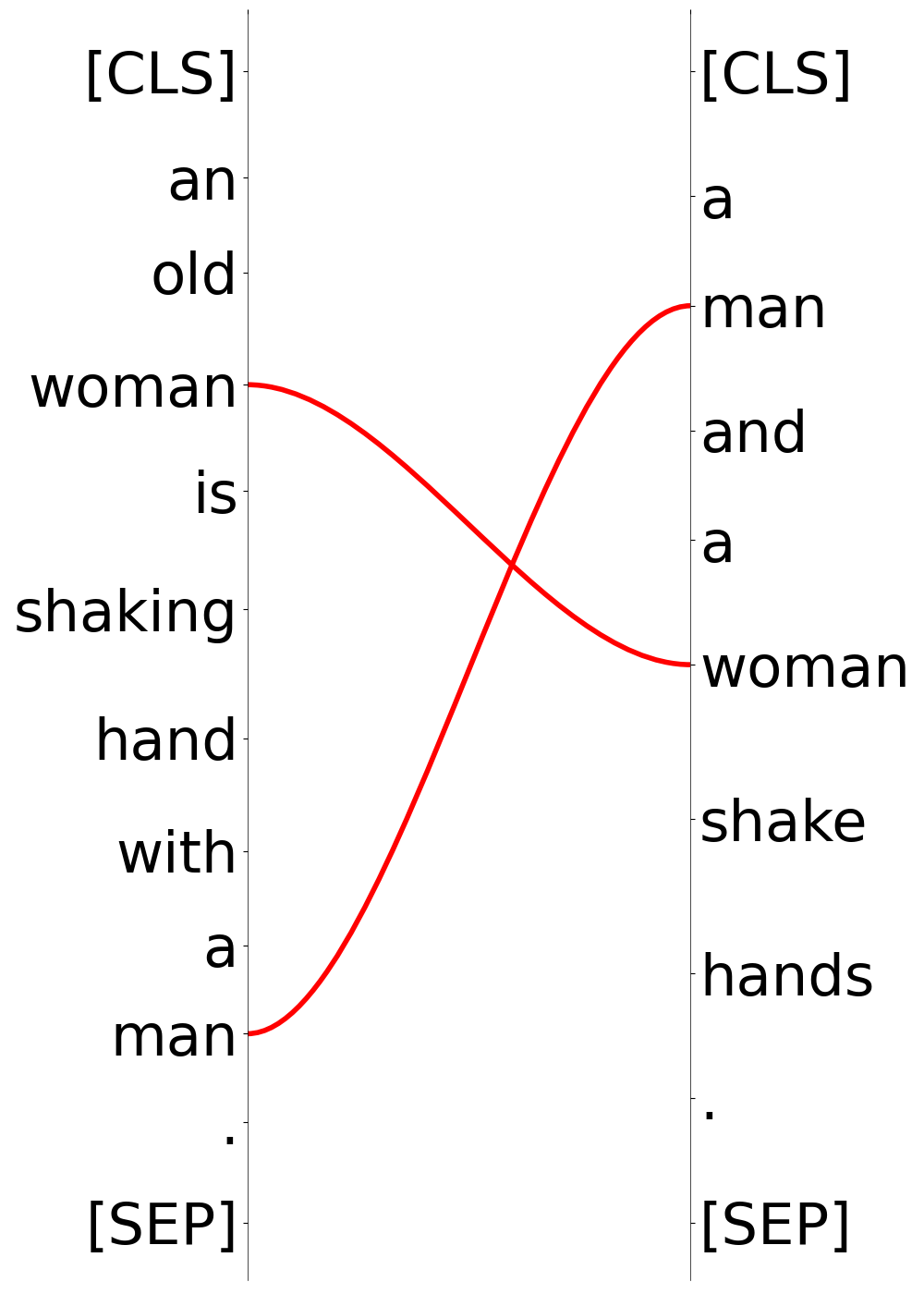}& \includegraphics[width=\linewidth]{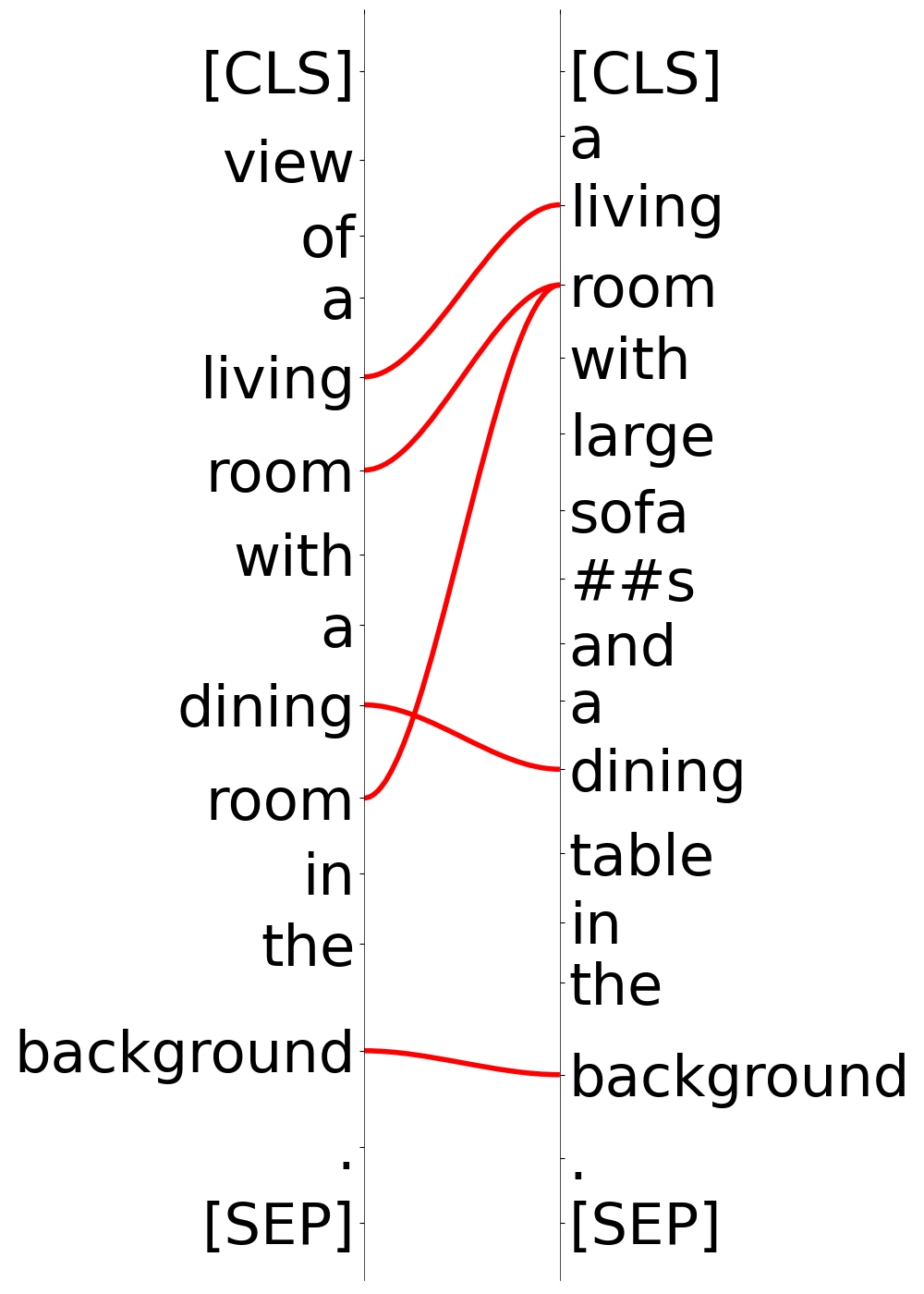}\\
     \SetCell[c=1]{l} Embeddings & &  \hfill 0.67\\
     \includegraphics[width=\linewidth]{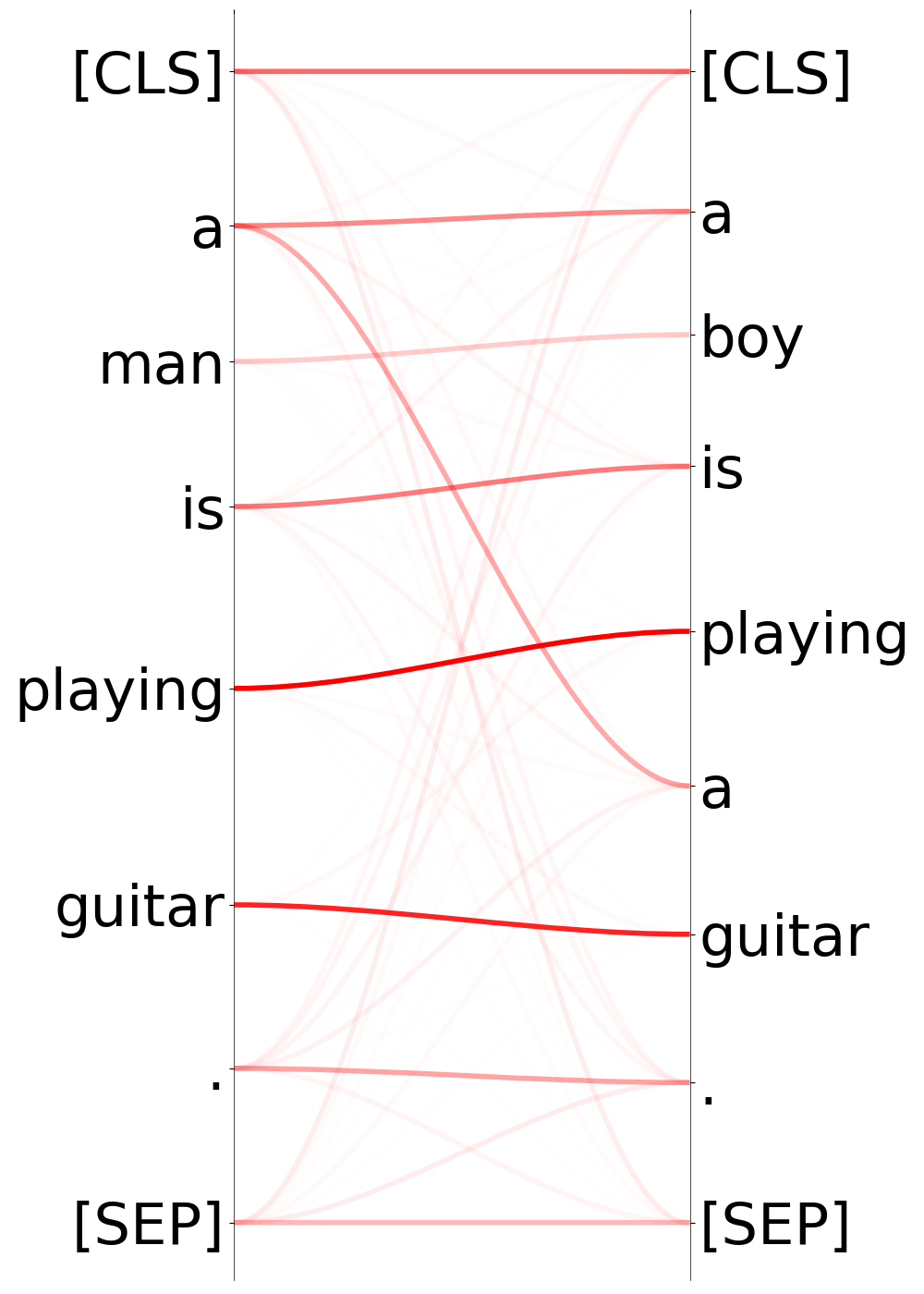} & 
     \includegraphics[width=\linewidth]{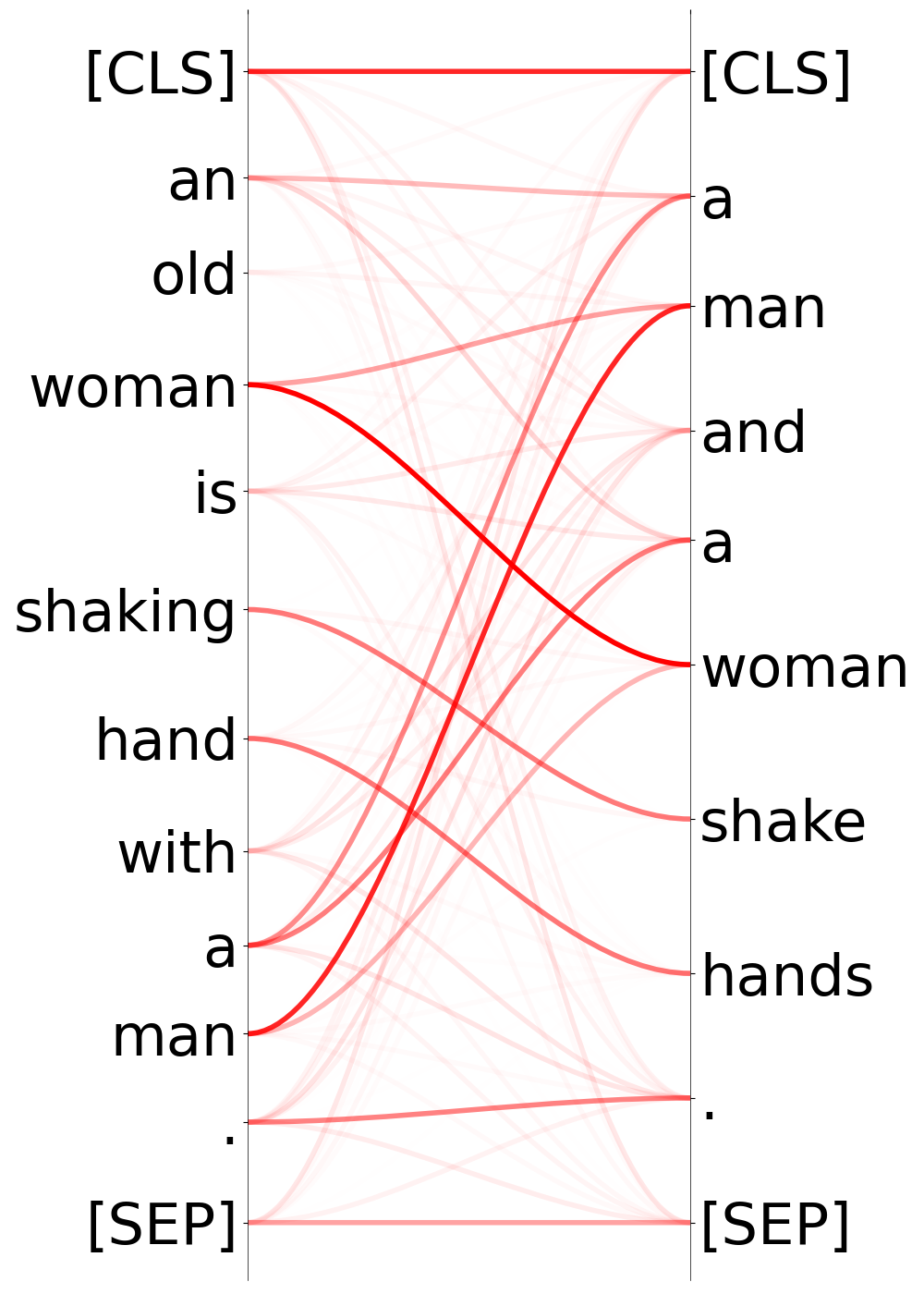} & \includegraphics[width=\linewidth]{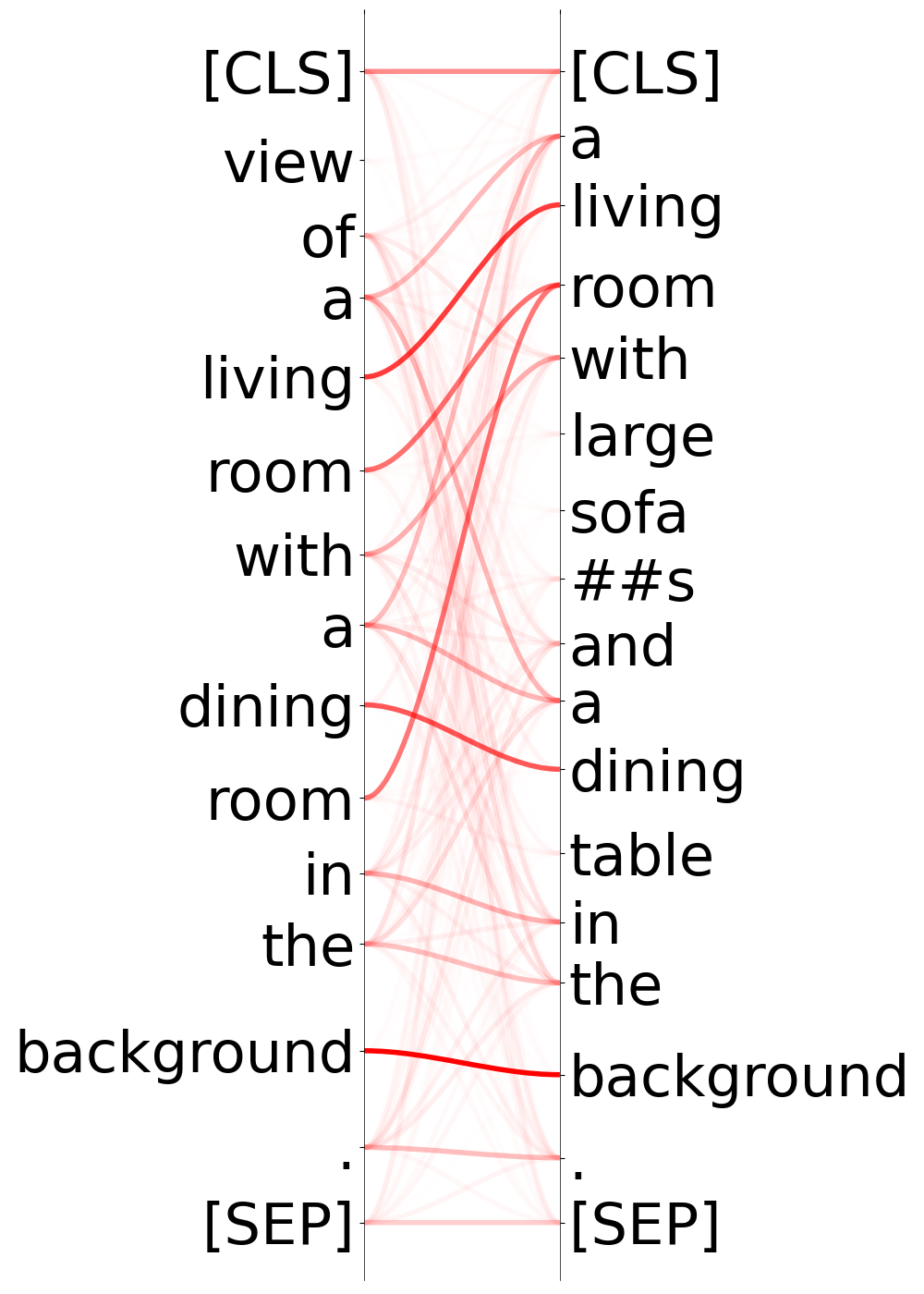}\\
       \SetCell[c=1]{l} HxP & &  \hfill 0.62\\
      \includegraphics[width=\linewidth]{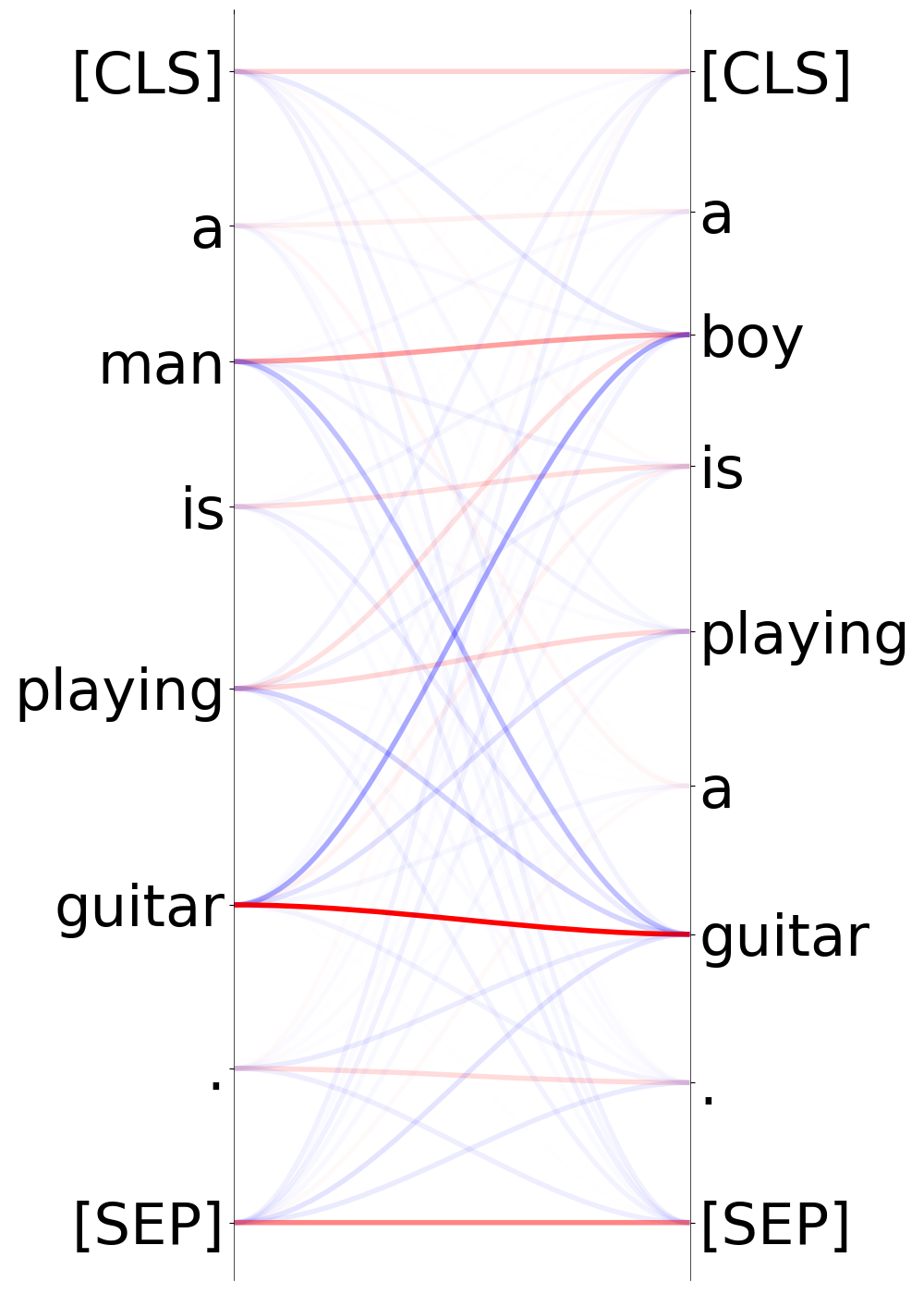} & 
       \includegraphics[width=\linewidth]{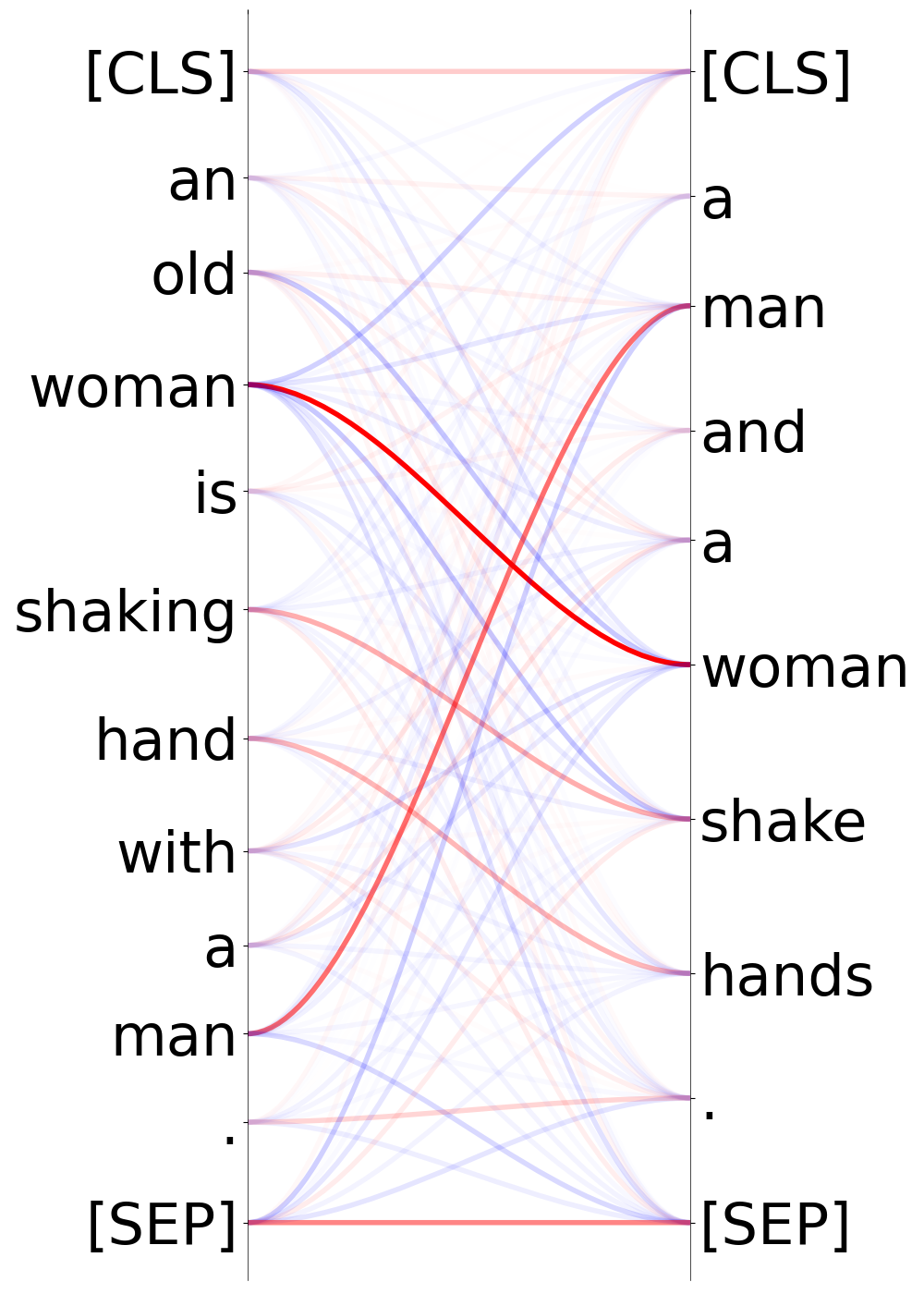} & \includegraphics[width=\linewidth]{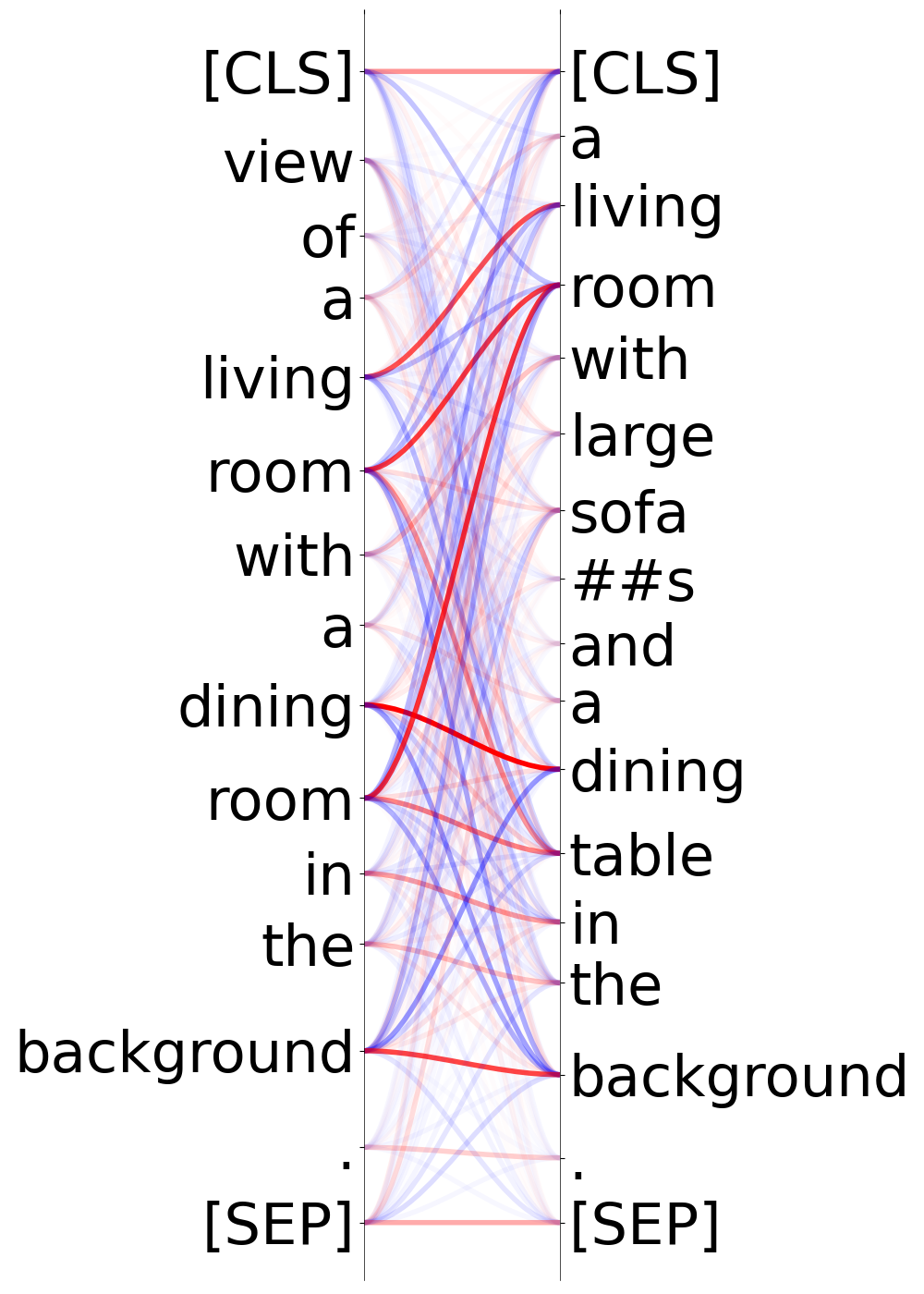}\\
      \SetCell[c=1]{l} BiLRP & &  \hfill \textbf{0.81}\\
     \includegraphics[width=\linewidth]{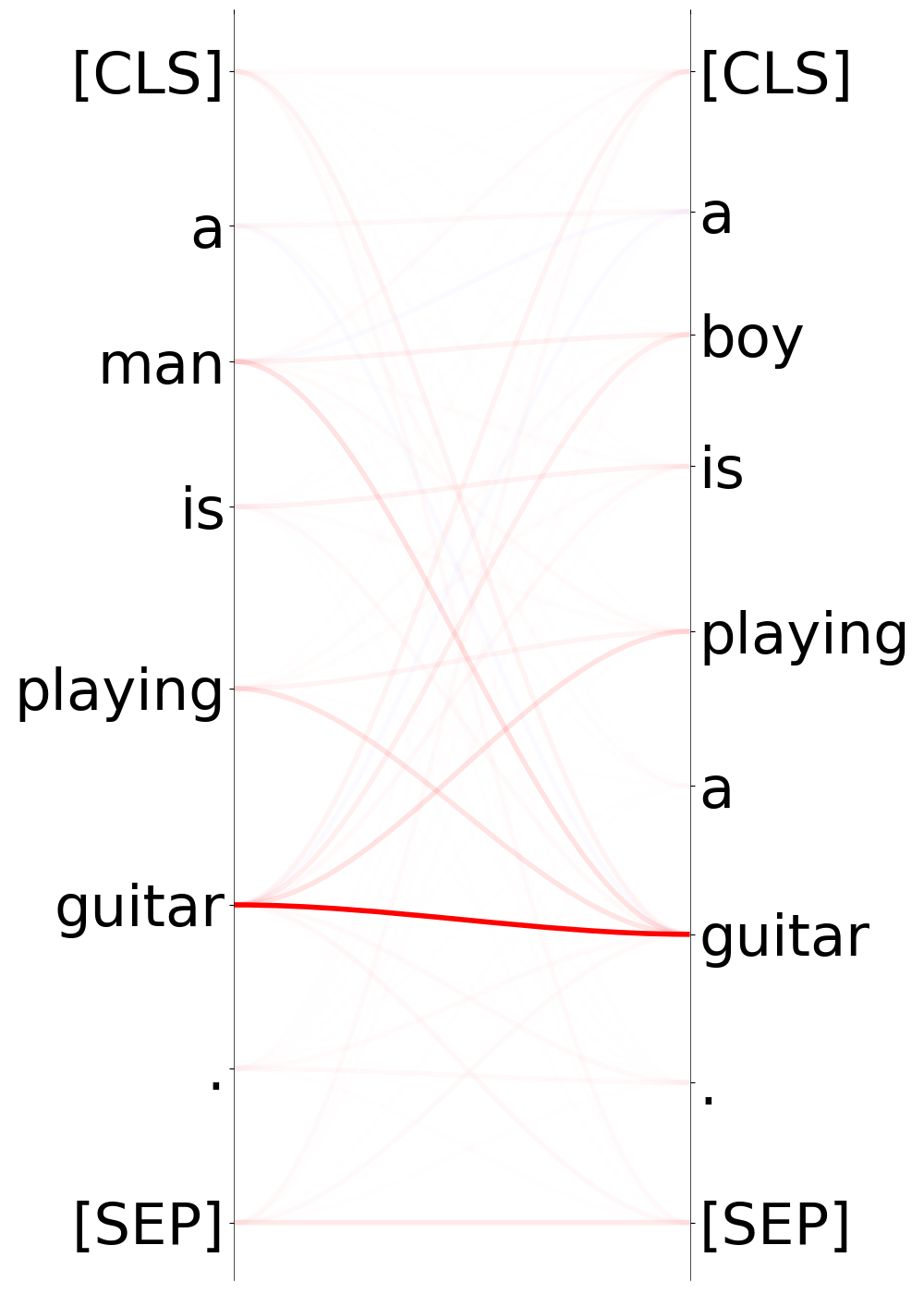} & 
     \includegraphics[width=\linewidth]{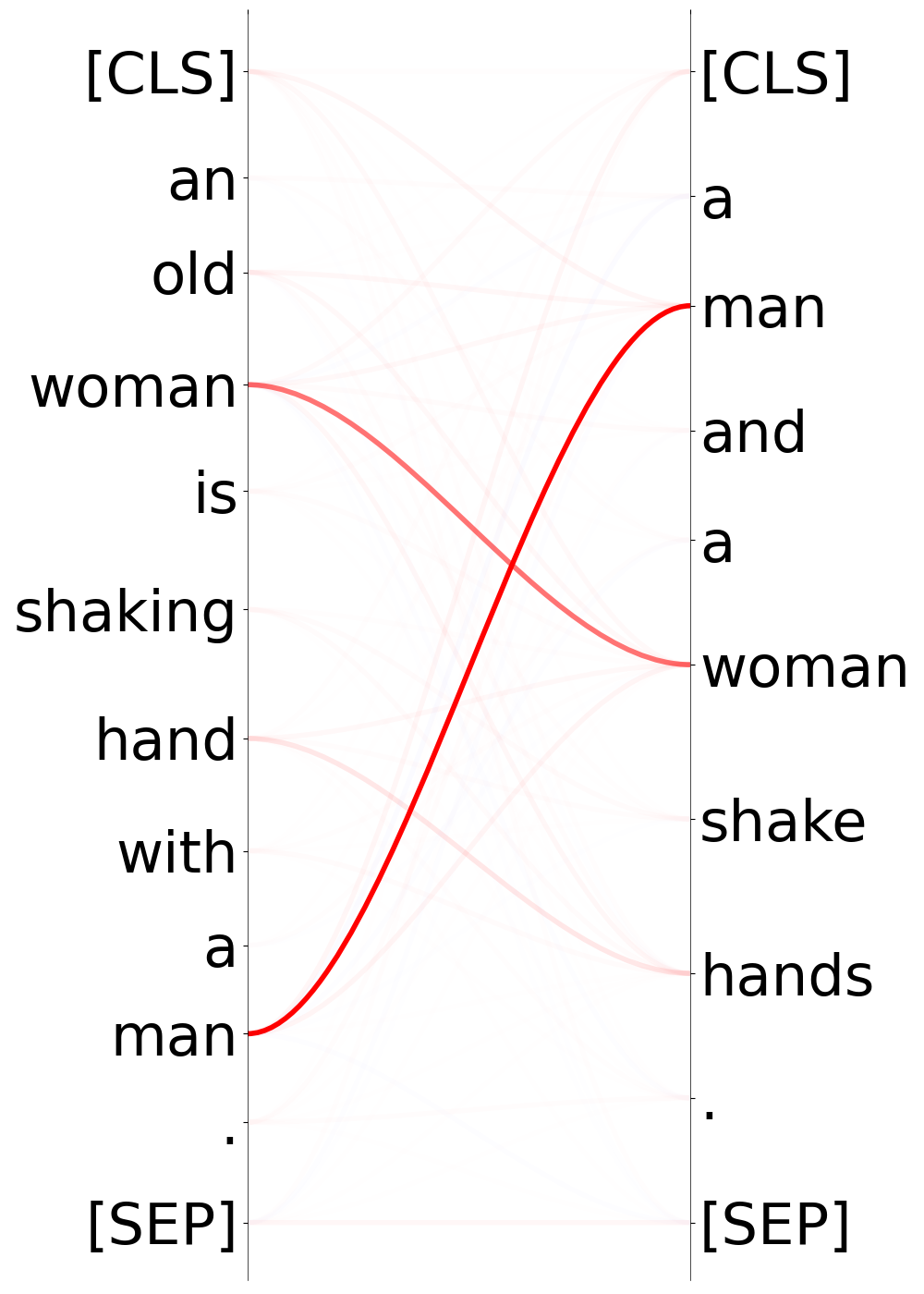} & \includegraphics[width=\linewidth]{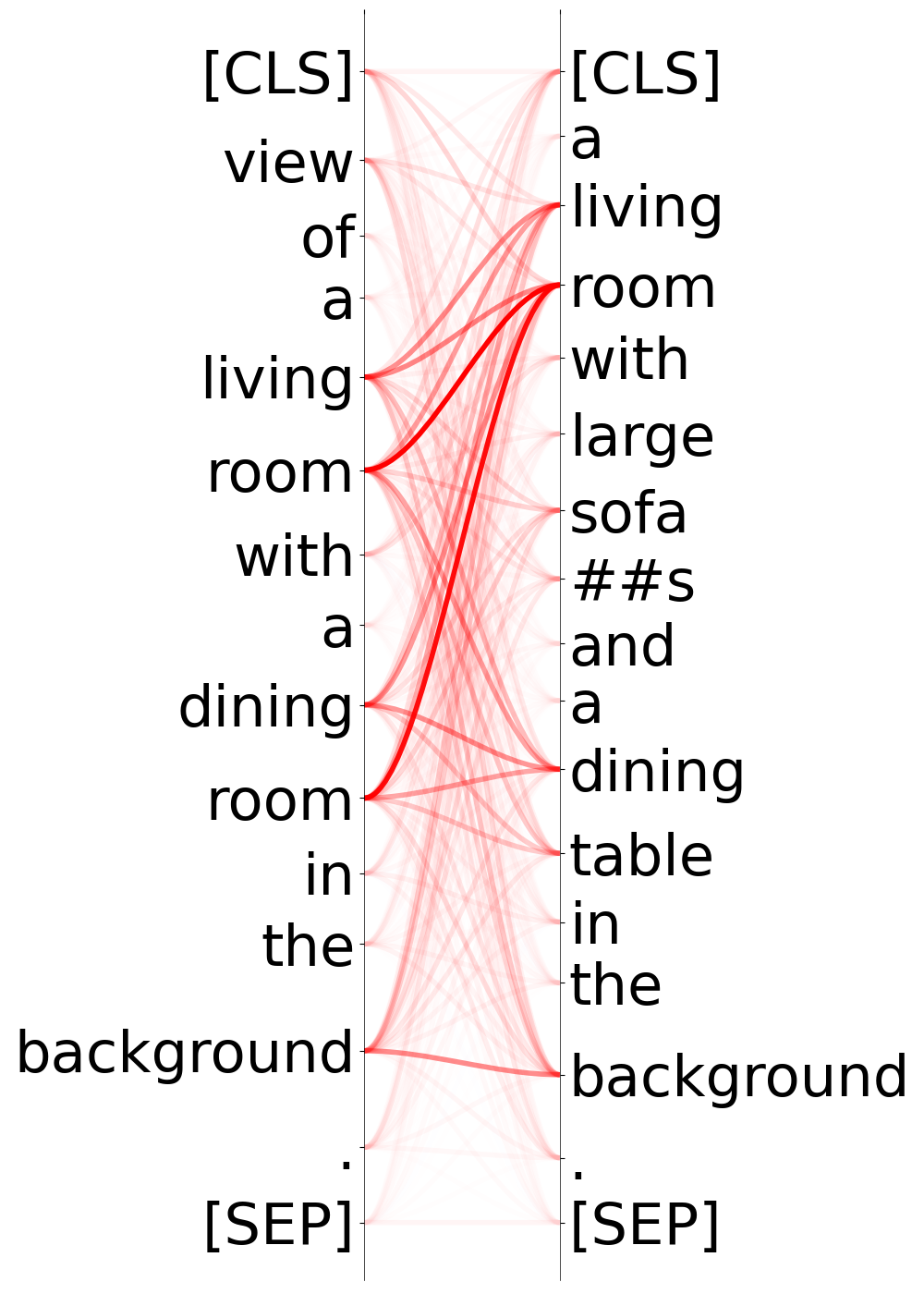}\\
\end{tblr}
\caption{Comparison of different explanation techniques that highlight the interaction between input features. Ground truth interactions (top row) are the interactions between same noun tokens. These are compared to second-order explanations built on top of BERT token embeddings, Hessian$\times$Product (H$\times$P) and BiLRP. Average cosine similarity (ACS) is used to measure agreement between ground truth and explanations. }
\label{fig:toy_evaluation}
\vspace{-2mm}
\end{figure}

\paragraph{Interaction Analysis}
The complex structure of the similarity task and a lack of fine-grained ground truth rationales of feature interaction motivates the evaluation of explanations on a specifically designed similarity task. We design a similarity model based on co-occurrence statistics of features, which here are interactions between same noun tokens (see Figure \ref{fig:toy_evaluation}, top row).
We finetune a similarity model using a  BERT-base encoder with a QKV-pooling layer to correctly predict the number of co-occurring proper nouns and nouns in the STSb dataset. After optimization of the mean squared error (MSE) loss between true and predicted scores, the similarity model is able to correctly predict the number of interactions (Spearman's $\rho\,{=}\,0.94/0.89$, $\text{MSE}\,{=}\,0.21/0.87$ for train/test). To verify that predictions are built from the expected interactions, we compute the similarity between ground-truth interactions and compare them to the second-order explanations extracted from (i) token embeddings, (ii) Hessian$\,\times\,$Product (H$\times$P), and (iii) BiLRP. \par

In Figure \ref{fig:toy_evaluation}, we observe that computing interactions directly from token embeddings results in pairwise attributions mainly between same tokens that are not selective with respect to their assigned part-of-speech (POS) tag. 
For H$\times$P, the interactions are much more selective with regard to nouns and proper nouns, and we observe considerable token interactions that are assigned negative relevance. In the case of many interacting tokens that drive high similarity (Figure \ref{fig:toy_evaluation}, right column), we observe that it becomes increasingly difficult to identify the relevant interactions. BiLRP is able to select the relevant tokens in comparison to the other baselines with higher accuracy, which is supported by the highest average cosine similarity (ACS) between true interactions and BiLRP of 0.81 in comparison to 0.62 for H$\times$P, and 0.67 for the embedding baseline.

In summary, our self-designed similarity task can be accurately explained using BiLRP, allowing a better understanding of the model's strategy of filtering out task-irrelevant parts of speech to solve the noun-matching task.

\begin{figure}
  \begin{subfigure}{\linewidth}
    \centering \scriptsize STSb
    \includegraphics[width=1.03\linewidth]{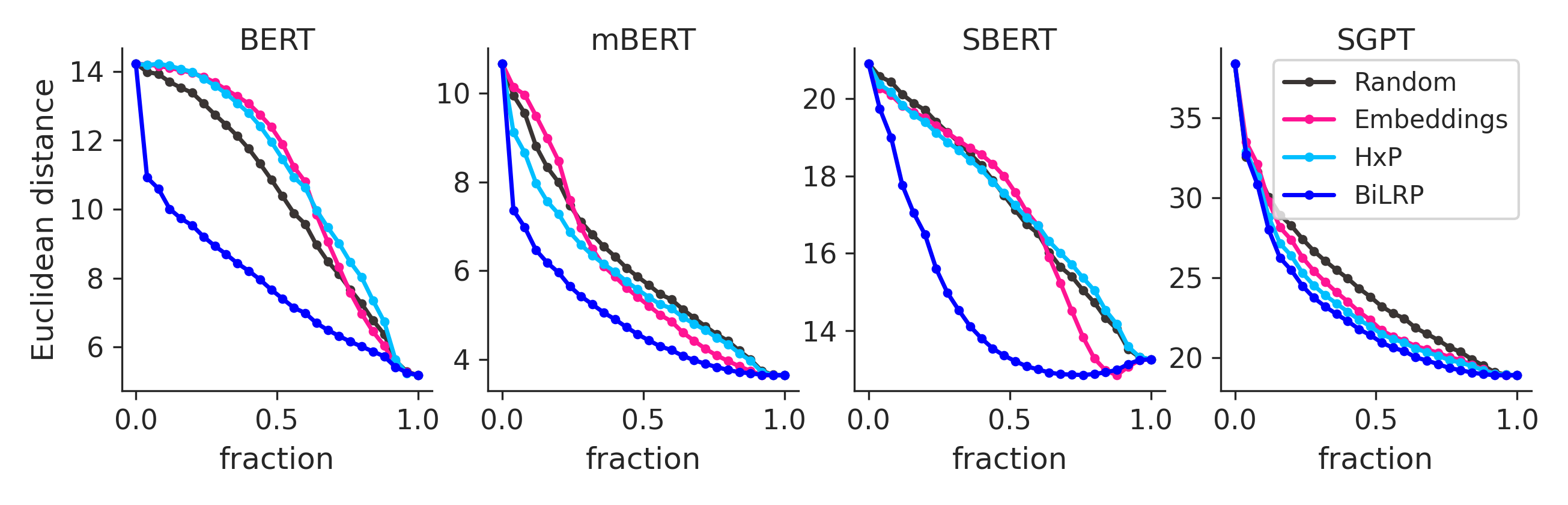}
  \end{subfigure}
  
  \begin{subfigure}{\linewidth}
    \centering \scriptsize BIOSSES
    \includegraphics[width=1.03\linewidth]{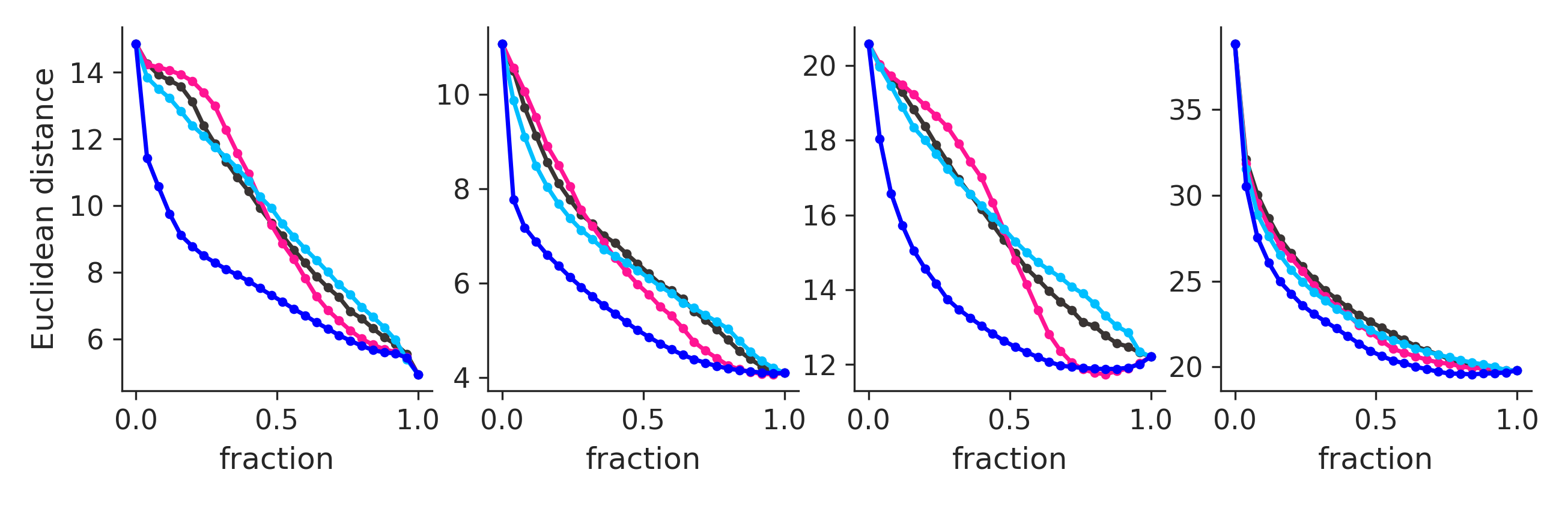}
  \end{subfigure}

  \caption{Perturbation experiment comparing different explanation methods across models. Fractions of tokens, ranked from most to least relevant, are added to one input sequence and the resulting Euclidean distance to the unperturbed sentence is measured. A steep initial decline with a smaller area under the curve indicates better identification of task-relevant features.
  }
    \normalsize
  \label{figure:perturbation}
  \vspace{-2mm}
\end{figure}

\paragraph{Perturbation Analysis}
We further test the ability of explanations to faithfully capture the similarity prediction process on real-world semantic similarity sentence pairs. Sequence elements are ordered based on sum-pooled interaction scores from highest to lowest relevance and elements are added iteratively to the selected input sequence. At each step, we compute the Euclidean distance between the perturbed and the unperturbed sentence representation, measuring how strongly the representation is affected in response to the removal of the next most relevant tokens. Resulting perturbation curves are shown in Figure \ref{figure:perturbation} comparing different explanation methods and a random baseline.
We observe that across models and datasets, BiLRP consistently selects the features that decrease the distance between sentence representations most effectively, resulting in the lowest area under the perturbation curve (see Appendix \ref{app:perturbation}). The steep initial decline in Euclidean distance for BiLRP  highlights that a small subset of highly relevant features are identified reliably. 
These findings further accentuate the differences between explanation methods, emphasizing the effectiveness of BiLRP in identifying the relevant interactions between tokens.

\paragraph{Conservation} The main axiomatic principle used to develop improved BiLRP explanations for Transformers is conservation of relevance. The sum of observed relevance is  directly related to the explained model predictions, and explanations that are relevance conserving have been shown to result in improved explanations \cite{transformerlrp2022}, as also confirmed in our experiments presented in Figures \ref{fig:toy_evaluation} and \ref{figure:perturbation}. Implementing these rules to compute second-order explanations (see Section \ref{sec:propagation_rules}) reconstitutes relevance conservation in BiLRP in comparison to H$\times$P as shown in Figure \ref{fig:conservation} in Appendix \ref{app:conservation}. A residual lack of conservation can be explained by neuron biases, which are unattributable (Figure \ref{fig:conservation}, rightmost panel). \par

Overall, these experiments have confirmed that BiLRP explanations can accurately identify the most relevant feature interactions, leading us next to an evaluation of the textual similarity prediction task itself. 

\subsection{Predicting Semantic Textual Similarity} \label{subsec:correlation}

Next, we focus on the task of predicting semantic textual similarity and compare different pre-trained similarity models (BERT, mBERT, SGPT, SBERT) across three datasets (STSb, SICK, BIOSSES). We evaluate model predictions using Spearman correlation with the goal to identify performance differences between models that inform more focused analyses using explainable AI.

\begin{table}[ht]
\centering
\scriptsize
\begin{tabular}{l@{~}ccc}
       \toprule
         Model & \multicolumn{3}{c}{Spearman correlation}  \\
          & STSb   &  SICK  & BIOSSES \\
       \midrule 
         BERT + CLS  & 20.3 &  42.4  & 63.7 \\
         BERT + Mean Pooling & 47.3 &  58.2  & 54.6\\
         mBERT +  Mean Pooling & 55.2 & 56.3  & 55.6\\
         SGPT + Mean Pooling & 76.9 &  73.4 & 69.6\\
         SBERT + Mean Pooling & 84.7 &  78.4 & 66.7 \\
         \bottomrule
\end{tabular}
\caption{Spearman correlation $\rho \times 100$ for a set of tasks and models. The first three models were not finetuned,  while SGPT and SBERT were finetuned on semantic similarity data.}
\label{table:stsb_scores}
\vspace{-2mm}
\end{table}

As shown in Table \ref{table:stsb_scores}, the standard BERT similarity model, when used with CLS or mean pooling methods, fails to effectively capture semantic proximity. The Spearman correlation $\rho\times 100$ for the CLS-Pooling is 20.3, whereas a significantly improved correlation  is achieved when using mean pooling across all encoded token representations for STSb and SICK data. Similarity is best predicted by SGPT and SBERT with scores ranging from 66.7 to 84.7, highlighting overall the considerable impact of model selection, pooling strategy and dataset on task performance.\par

\section{Corpus-Level Use Cases}
To go beyond the mere evaluation of nominal correlation, which is at risk of obfuscating undesired model strategies, i.e. Clever-Hans-type and shortcut learning \cite{lapuschkin2019unmasking, Geirhosetal20}, we hereby explore how BiLRP explanations can be used to uncover general model strategies in three distinct use cases.\par

\subsection{Explaining Semantic Textual Similarity}
To conduct an explanation-based analysis of semantic similarity, we retrieve explanations for all 1379 samples of the STSb test set. 
Token-to-token interactions are summarized by extracting POS tags using \textit{spaCy 3} and aligning different tokenizers using \textit{tokenizations} \footnote{\url{https://github.com/explosion/tokenizations}}. 

The corpus-level analysis is performed by aggregating all relevance scores per token pair. Relevance for each interaction between POS-tags is aggregated and scores are normalized  by the maximum absolute value of total summed relevance, which results in a relevance scoring over interactions, as shown in Figure \ref{fig:corpus_level}. For each interaction, we distinguish between relevance patterns that negatively (blue triangle) or positively (red triangle) contribute to the similarity score.

For the BERT + CLS similarity model with lowest correlation scores as shown in Section \ref{subsec:correlation}, we identify that the most positively relevant interactions are  `NOUN-NOUN', `NOUN-VERB' and `NOUN-[SEP]' (Figure \ref{fig:corpus_level}-a). While negative contributions are less pronounced, we observe some amount of negatively contributing `NOUN-[CLS]' interaction. For mean pooling shown in Figure \ref{fig:corpus_level}-b, we observe the strongest positive effects for `[SEP]-[SEP]', followed by `NOUN-NOUN' and `NOUN-VERB' interactions. Overall, the distribution of relevance is more concentrated over a smaller set of POS interactions. 
The SBERT model shows a distribution comparable to BERT + CLS, focusing on a similar subset of interactions (Figure \ref{fig:corpus_level}-c), though assigning different importance to them. Specifically, `NOUN-NOUN', `NOUN-[SEP]' and `NOUN-VERB' are most relevant. In contrast to BERT with mean pooling, the `[SEP]-[SEP]' interaction is significantly less pronounced. SBERT also attributes considerable amount of relevance to `VERB-VERB' and `NOUN-ADJECTIVE' interactions, which overall suggests that the semantic similarity task can be solved quite well using a small but well-chosen subset of POS interactions.

Our analyses further provide insights regarding the choice of pooling strategy, supporting previous findings that are in favor of mean pooling over CLS pooling \cite{mohebbi-etal-2021-exploring} and underscoring the usefulness of explanations on the level of interactions to identify differing strategies across models.

\begin{table*}[t!]
\begin{subfigure}{0.33\textwidth}
\caption*{\textmyfont{\textbf{a.}} BERT + CLS}\vspace{-1mm}
\includegraphics[width=\textwidth]{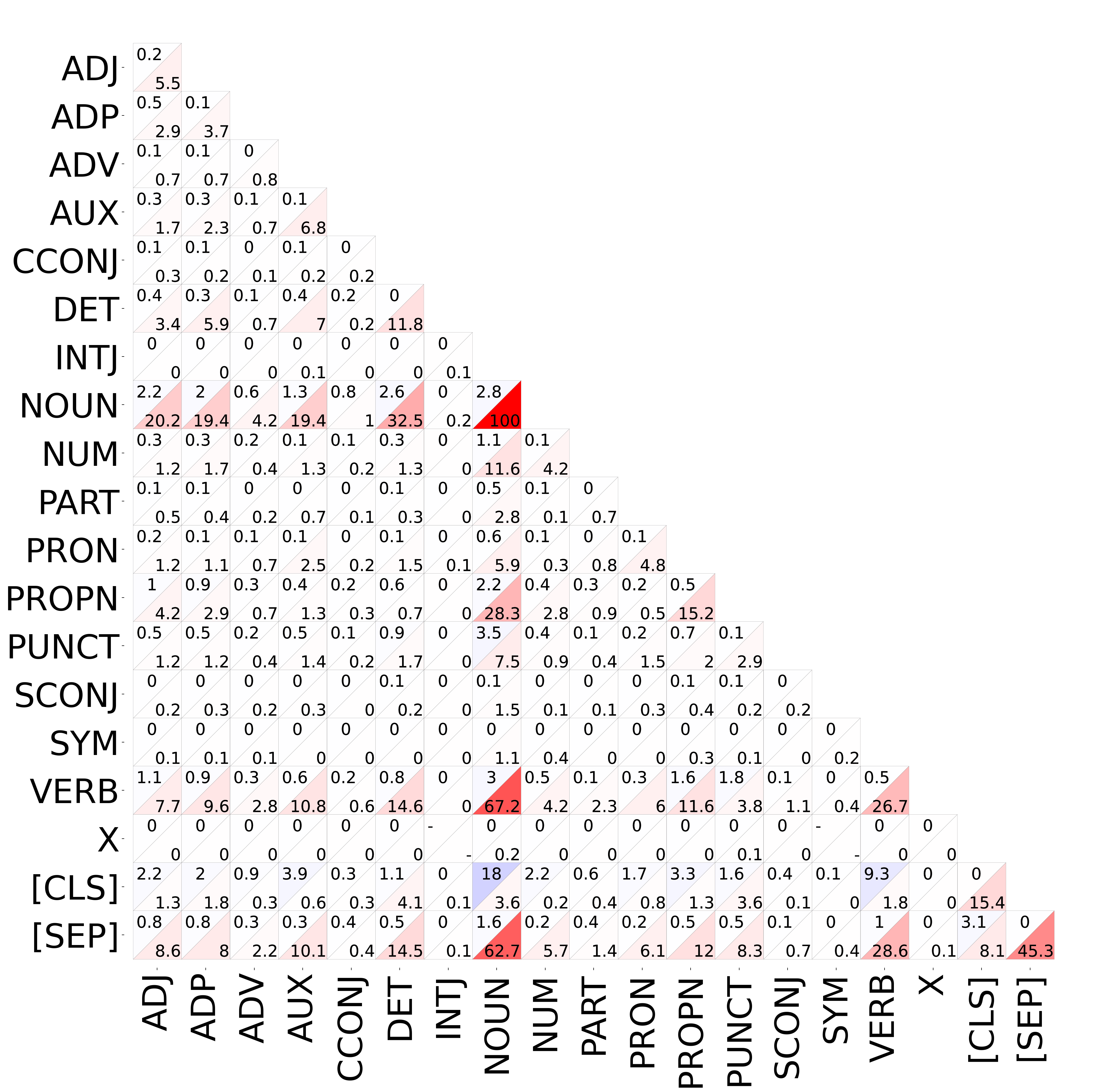}
\end{subfigure}\hfill
\begin{subfigure}{0.33\textwidth}
\caption*{\textmyfont{\textbf{b.}} BERT + Mean Pool.}\vspace{-1mm}
\includegraphics[width=\textwidth]{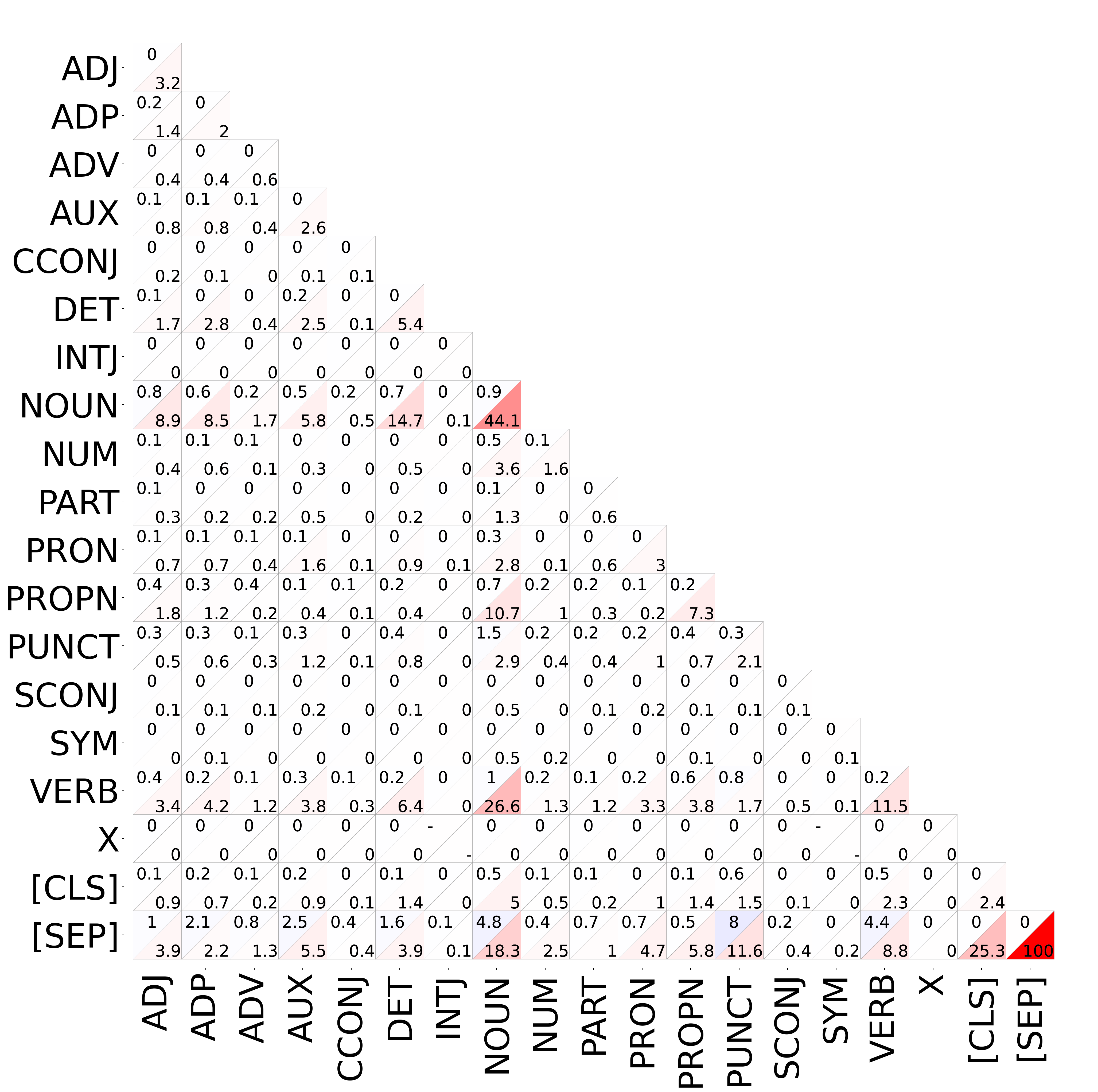}               
\end{subfigure}\hfill
\begin{subfigure}{0.33\textwidth}
\caption*{\textmyfont{\textbf{c.}} SBERT + Mean Pool.}\vspace{-1mm}
\includegraphics[width=\textwidth]{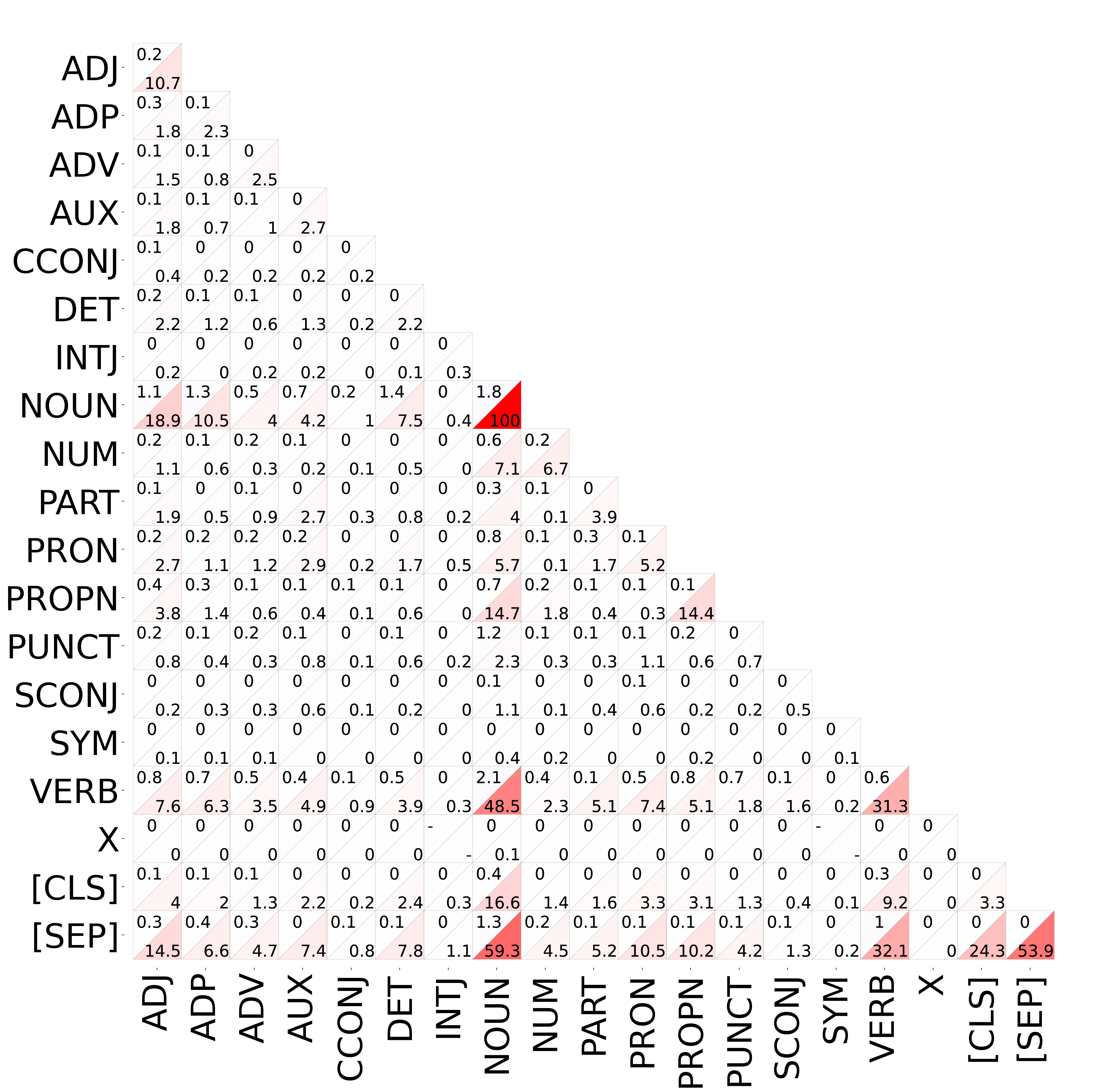}              
\end{subfigure}
\captionof{figure}{Corpus-level analysis of BiLRP explanations between POS tags on the STSb dataset. The contribution of positive/negative interactions to the similarity score is shown in red/blue for three similarity models, ranging from (\textbf{a}) the least predictive (BERT + CLS), to (\textbf{b}) moderately predictive (BERT + Mean Pooling), to (\textbf{c}) the most predictive (SBERT) (cf. Table \ref{table:stsb_scores}).  }
\addtocounter{table}{-1}
\label{fig:corpus_level}
\vspace{-2mm}
\end{table*}

\subsection{Explaining Multilingual Similarity} 
\label{sec:multilingual}

Multilingual language models enable the flexible use of embeddings  for unsupervised downstream tasks across different languages. In this section, we consider a setting in which a ranking of most similar texts in a multilingual database is required. 
We use the mSTSb dataset and compute similarity for mBERT and a set of monolingual BERT models for English, German, Russian and Chinese, assuming that no fine-tuning on semantic similarity is performed. The text representations are extracted by mean pooling.

\begin{figure}[h!]
    \centering\includegraphics[width=\linewidth]{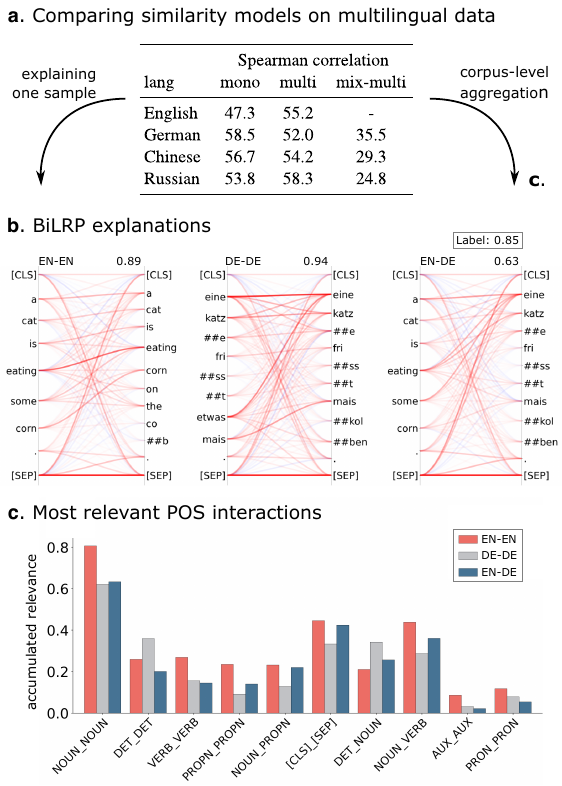}
    \vspace{-6mm}
    \caption{Comparison of mono- and multilingual BERT-based similarity models on mSTSb. (\textbf{a}) Spearman correlation $\rho \times 100$ of the multilingual STSb corpus. Similarity models are build from monolingual (mono) and multilingual (multi) BERT models that receive monolingual input, and a multilingual model that receives mixed input in English and a translated version of the other sentence (mix-multi).
    (\textbf{b}) BiLRP explanations on mBERT for English-English (left), German-German (center) and English-German (right). The sentence pair is assigned a true similarity score of 0.85.
    (\textbf{c}) Comparison of positively relevant POS interactions. POS tags are selected based on largest difference of accumulated relevance  between the mixed and the monolingual settings. 
    }
    \label{fig:multilingual_pos}
    \vspace{-1mm}
\end{figure}

\paragraph{Results}
We first analyze the correlation to ground truth similarity scores and focus on nominal prediction differences across settings, guiding the selection of an interesting case study for our subsequent BiLRP analysis. As shown in Figure \ref{fig:multilingual_pos}-a, we observe that Spearman correlation scores are consistently between 47.3 and 58.5 across the four aforementioned languages, reaching similar correlation levels as observed in the previous section for not finetuned similarity models (cf. Table \ref{table:stsb_scores}). Interestingly, for the mixed-multilingual case, we observe a clear drop of correlation scores to 24.8--35.5. To uncover some of the effects that drive this decrease in performance, we next conduct a case study on the English and German subsets of the mSTSb corpus.

In Figure \ref{fig:multilingual_pos}-b, our initial step involves explaining a sentence pair using BiLRP within monolingual settings (EN-EN, DE-DE)  and comparing it to the mixed setting  (EN-DE).
For English, we see how semantic similarity is attributed to the interaction of `eating’, whereas the German translation `frisst’ is considered less relevant. Instead, we find that semantic similarity in the German setting is more often attributed to the interaction between determiners (`eine-eine’), which may reflect the specific role of determiners in the German language that both quantify and determine an object \cite{dipper2005}. We provide additional samples that illustrate several cases of relevant interactions  in Appendix \ref{app:multilingual_triplets}. These include the mismatching of different quantities (`two-three’), effects of matching subtokens that affect the semantic meaning (`key-\#\#board’ and `keyboard’), and overall overconfidence of the model in assigning high similarity based on semantically related tokens (`train-waiting’, `clothing-shirt’). \par

We take a closer look at the aggregated most relevant POS interactions with the goal to identify differences in model strategies across  settings. Specifically, we analyze which POS interactions change the most from the monolingual to the multilingual setting and show the ten interactions of largest difference in accumulated positive relevance assigned to a specific POS interaction in Figure \ref{fig:multilingual_pos}-c. We additionally show negative changes in relevance in Figure \ref{app:fig:multilingual_neg} in Appendix \ref{app:multilingual_triplets}. We find that both `NOUN-NOUN'  and `VERB-VERB' interactions are less relevant in the mixed setting when compared to the monolingual English setting, suggesting that the model is less able to match English to German nouns and verbs respectively. Furthermore, we find that the monolingual German similarity is driven by a considerable amount of interaction between determiner tags (`DET-DET') that are less present in the mixed case. Lastly, we underline the difference in assigned relevance to the interaction between nouns and proper nouns (`NOUN-PROPN'). We hypothesize that, while multilingual models learn a flexible joint embedding space across many languages and vocabularies, it may be difficult to accurately model subtle differences in semantic meaning that monolingual models are able to capture more precisely.

These detailed insights, both at the single-sample and corpus levels, can unveil the diverse strategies that result in accurate or inaccurate predictions of similarity, potentially informing the future design of similarity tasks and models across multilingual settings.

\subsection{Analyzing Model Differences}
\begin{figure}[htbp]
  \centering
  \scriptsize
  \begin{minipage}[t]{0.52\linewidth}
    \centering
    \includegraphics[trim={10pt 0 0 0},clip, width=1.2\textwidth]{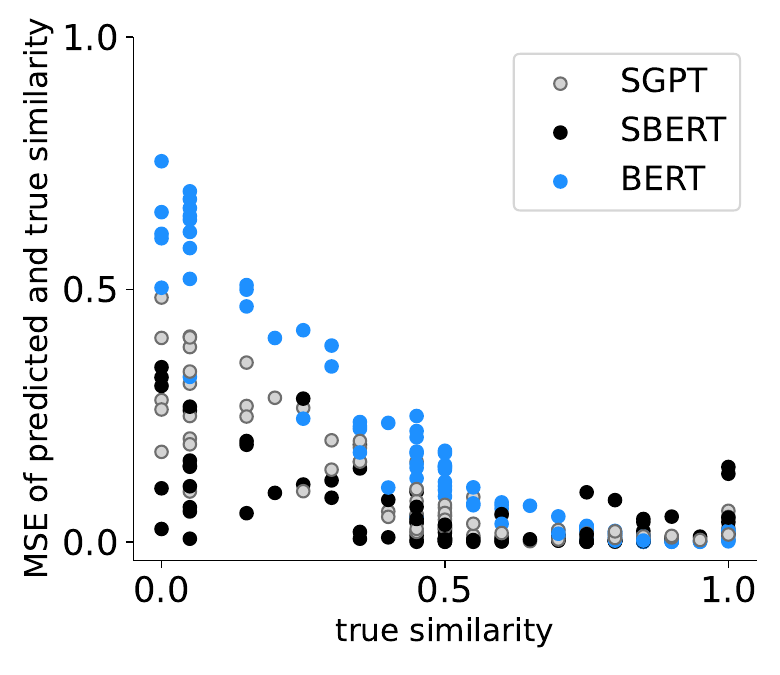}  \vspace{2pt}
  \end{minipage}
  \hfill
  \begin{minipage}[t]{\linewidth}
    \centering
    \begin{tblr}{
        colspec = {X[0.04]XXX},
        cells={halign=c, valign=m},
        rowsep=1pt, 
        colsep=1.pt, 
        leftsep=1pt, 
        rightsep=1pt, 
      }
       & \textbf{SGPT} & \textbf{SBERT} & \textbf{BERT (mean)}\\
      \hline
        \SetCell[r=5]{c} \rotatebox{90}{high} & miR-126--reports & up--up & tumor--tumor \\
        & miR-223--human & leukemia--leukemia & leukemia--leukemia \\
         & miR-223--regulates & mutual--mutually & cancers--cancers \\
        & regulated--of & also--up & cancer--cancer \\
        & of--cervical & cancer--cancer & tissues--cancer \\
       \hline
       \hline
        \SetCell[r=5]{c} \rotatebox{90}{low}  &  dependent--miR-133b & tumors--tumors & mutant--mutations \\
&  KRAS--driven & nucleotides--rna & dependent--mutations \\
&  oncogenic--NSCLCs & research--research & oncogenic--mutations \\
&  oxidative--downward & tumor--cancer & mutations--found \\
&  combined--viability & also--augmented & vivo--tumors \\
      \hline
    \end{tblr}
  \end{minipage}
\caption{Analysis of semantic similarity on the BIOSSES dataset containing biomedical text. Mean squared error (MSE) between predicted and true similarity for SGPT, SBERT and BERT similarity model is shown (top). Top-5 most relevant token interactions are shown for high and low similarity levels (bottom).}
\label{fig:biosess_analysis}
\vspace{-1mm}
\end{figure}

In the next use case, we analyze what drives performance differences observed earlier in Table \ref{table:stsb_scores} on the biomedical text pairs contained in BIOSSES \cite{biossess2017}.
In a first step, we analyze the predicted similarity scores in Figure \ref{fig:biosess_analysis} (top) for SGPT, SBERT and BERT as introduced in Section \ref{subsec:correlation}. We clearly see how the error between ground truth and predicted similarity decreases for higher levels of true similarity. All three models assign correct levels of similarity for high similarity but are less capable of correctly capturing low similarity. \par

In the second step, we compute explanations and select the 25\% data quantiles of highest and lowest predicted similarity. For each group, we rank the most relevant interactions and show the top-5 for each model in Figure 
\ref{fig:biosess_analysis} (bottom). For high similarity, we observe that both SBERT and BERT base their similarity predictions primarily on matching of same tokens or variants thereof, e.g. `leukemia-leukemia' and `cancer-cancers', while SGPT explanations identify interactions with gene regulating molecules (`miR-126-reports' and `miR-223-regulates'), and interactions to more functional token pairs (`regulated-of') as most relevant. Similarly, for low similarity,  SGPT explanations assign high relevance to molecule-specific interactions (`dependent-miR-133b' and `KRAS-driven') and more general biomedical vocabulary (`oxidative--downward'). SBERT and BERT both show a comparable matching pattern (`also-augmented', `mutations-found'), but as for high similarity also match closely related biomedical tokens (`nucleotides-rna', `tumor-cancer').

This suggests that SGPT has better abilities to match task-relevant biomedical vocabulary,  specifically names of gene molecules and tokens that are descriptive of gene expression processes, while SBERT und BERT base their similarity predictions on more general medical terminology. These strategies work well for highly similar sentence pairs for which matching may be sufficient, but are less suitable to correctly assess low similarity that may require more complex strategies e.g. the identification of negations or counterfactuals. 

\section{Discussion and Conclusion}
Our evaluation and use cases have provided a framework to analyze textual similarity models using explainable AI. In the following, we contextualize our findings and discuss further implications.

\paragraph{Structured Explanations}
We have seen how novel types of explanations can be used to explain bilinear models in the context of semantic textual similarity. Resulting explanations highlight pairs of features that are most or least relevant to produce a particular similarity score. These second-order attributions provide an appropriate level of complexity and detail for investigating similarity model predictions that go beyond heatmap representations. 
In our experiments, BiLRP identifies feature interactions in Transformers more accurately than the token embedding baseline and the Hessian-based H$\times$P explanations. This has allowed us to analyze the unsupervised semantic similarity task on the level of fine-grained interactions. Our results have highlighted interpretable interactions of tokens and POS tags that can next be utilized to inform and guide the development of similarity tasks and models. 
In particular, leveraging generative approaches may further inform the development of interaction-based explanation techniques.

\paragraph{Strategies of Similarity Models}
This approach enables insights into the internal computation of sentence representations in Transformers, exposing unexpected strategies employed for predicting semantic similarity. In particular, we have observed that high relevance in non-finetuned similarity models is often assigned to interactions between same tokens, revealing a rather simple token matching strategy. Our corpus-level POS analysis has indicated that semantic similarity can be approached by a small subset of interactions, most importantly interactions between nouns or proper nouns, verbs, and nouns and verbs, which promotes the notion of similarity as a `bag of interactions'. We observe high relevance in specific token interactions (CLS and SEP), consistent with prior research \cite{clark-etal-2019-bert, bondarenko2023quantizable}. In extension, this may result in unintended matching of tokens that is semantically not meaningful. Our experiments suggest that fine-tuning on similarity tasks and selecting an appropriate pooling strategy can partially alleviate these effects. While our focus has been on models predicting similarity, these structures are also commonly used as part of internal computations, such as matching key and value representations, which could also be analyzed using interactions.

\paragraph{Conclusion}
In many language applications the computation of similarity is a central concept to match  textual information. Here, we have shown how predictions of widely used deep similarity models can be analyzed by assigning relevance to the interaction of features. These explanations have offered novel ways to perform corpus-wide analyses ranging from information retrieval, to multilingual text matching, and the estimation of similarity in biomedical text data.

Considering the fast-growing number of applications based on foundation models, we believe that explainable AI has a critical role in ensuring their safe, robust and compliant use.
AI safety is especially critical in high-risk domains including legal, financial, medical, or governmental applications of machine learning for which text data plays a central role, accentuating the necessity for rigorous system evaluations and comprehensive explanations.  
For complex tasks like semantic similarity, datasets that contain detailed structured rationales on the level of interactions are needed to improve model alignment with human expectations.

\section*{Limitations}

\paragraph{Methods} In this paper, we have focused on using post-hoc explanations and in particular gradient-based explanations, aiming for the most faithful explanations as identified by our evaluation. Achieving high faithfulness and meeting the conservation principle, requires implementation of gradient propagation rules, especially for a diverse range of Transformer architectures. 
Computing second-order explanations requires to compute as many backward passes as there are sentence embedding dimensions for each sentence in a pair. For the here considered embedding dimension of 768, this required around two minutes computation time on a 12GB P100 GPU. To accelerate computations, explanations can be computed in batch, provided that the memory requirements are met. The factorization of BiLRP enables the reuse of computed explanations when the same sentence or pair reoccurs in subsequent instances.

\paragraph{Evaluation}
Regarding our evaluation, we have adapted standard perturbation experiments designed for evaluating first-order heatmaps to second-order explanations. We did this by initially sum-pooling over one of the sentences, performing perturbations on the corresponding input sequence, and subsequently repeating the analysis for the other sentence. This allowed us to measure how well the identified explanations are able to affect the closeness of the sentence representations, measuring faithfulness of our explanations with respect to the predictions. To evaluate the ability of second-order explanations in detecting relevant interactions, we have designed a synthetic experiment. While this does not cover the semantic complexity of real-world semantics, we consider it an important step to motivate the collection of structured rationales for complex language tasks.

\paragraph{Analyses}  We have focused our analyses on a few selected datasets but expect that our insights, i.e. regarding token matching, apply to the task of semantic similarity more broadly across different models and corpora.
Analysing the effect of POS tags removes complex grammatical structures, e.g. verbal or gerundial nouns, which would furthermore complicate our multilingual analyses due to diverging grammatical rules and a varying number of categories.  Thus, we consider the focus on POS-level granularity to be an appropriate first step at an intermediate level of detail to investigate the task of semantic similarity at a corpus scale. 

\section*{Ethics Statement}
We do not anticipate any harmful risks in using the methods and analysis used in this work. 

\section*{Acknowledgements}
We would like to thank Klaus-Robert  M{\"{u}}ller for hosting AV as a student in his group, and providing valuable feedback on the manuscript. We also thank AM for proofreading. OE received funding by the German Ministry for Education and Research (under refs 01IS18056A and 01IS18025A) and BIFOLD.

\bibliography{anthology,custom}

\appendix

\section{Model details} \label{app:model_details}
We provide details regarding model type, the Hugging Face identifier and the implemented propagation rules for each model in Table  \ref{tab:model_details}. 

\section{Perturbation Details} \label{app:perturbation}

To compute the perturbation evaluation, for each input pair, we order sequence elements based on sum-pooled relevance interaction scores from highest to lowest relevance and add elements iteratively to the selected input sequence. Tokens are added in steps of fractions of 0.04 until the original sentence is fully recovered. The initial empty sequence is initialized from the masking token for BERT-based models and `Ġ' for SGPT since no specific masking or unknown special tokens are reserved by the tokenizer.
Perturbation curves are computed once for each sentence in a pair and the resulting scores are averaged over all samples  (1379 pairs of STSb and 100 pairs for SICK respectively).
In Table \ref{table:stsb_perturbation_scores_2} we report area under perturbation curve (AUPC) scores for the evaluation presented in the main paper Section \ref{sec:evaluation}. We further test if BiLRP AUPC scores are significantly lower than any of the other comparison methods and find this to be the case with $p\leq0.05$ for all investigated models and datasets. 

\begin{table*}
\centering
\scriptsize
\begin{tabular}{r l c }
\toprule
model & name & propagation rules\\
\midrule
BERT & bert-base-uncased & \multirow{ 2}{*}{AH, LN, GA} \\
mBERT & bert-base-multilinugal-uncased &  \\
\midrule
SBERT & sentence-transformers/stsb-bert-base & AH, LN, GA \\
SGPT & Muennighoff/SGPT-125M-mean-nli-bitfit & AH, LN \\
\midrule
German BERT & bert-base-german-cased & \multirow{ 3}{*}{AH, LN, GA} \\
Russian BERT & DeepPavlov/bert-base-bg-cs-pl-ru-cased &  \\
Chinese BERT & bert-base-chinese &  \\
\bottomrule
\end{tabular}
\caption{Used models and their Hugging Face identifier names alongside the used propagation rules. Attention Head (AH), Layer Normalization (LN) and GeLU Activation (GA) rules. }
\label{tab:model_details}
\end{table*}

\begin{table*}
\centering
\scriptsize
\begin{tabular}{lrrrr}
\toprule
 & Random & Embed. & \multicolumn{1}{c}{HxP} & \multicolumn{1}{c}{BiLRP} \\
\midrule
 &  \multicolumn{4}{c}{STSb (N=1379)} \\
\midrule
BERT & 10.28$\pm$1.64 & 10.86$\pm$1.15 & 10.97$\pm$1.50 & 7.80$\pm$0.94 \\
mBERT & 6.16$\pm$0.78 & 5.98$\pm$0.57 & 5.82$\pm$0.93 & 4.92$\pm$0.67 \\
SBERT & 17.19$\pm$2.60 & 16.94$\pm$2.26 & 17.20$\pm$2.58 & 14.52$\pm$3.41 \\
SGPT & 24.41$\pm$4.42 & 23.65$\pm$4.78 & 23.22$\pm$4.79 & 22.70$\pm$4.83 \\
\toprule
 &  \multicolumn{4}{c}{BIOSSES  (N=100)} \\
\midrule
BERT & 9.63$\pm$1.35 & 9.66$\pm$0.99 & 9.72$\pm$1.41 & 7.55$\pm$0.82 \\
mBERT & 6.58$\pm$0.81 & 6.38$\pm$0.54 & 6.42$\pm$1.06 & 5.34$\pm$0.51 \\
SBERT & 15.59$\pm$1.54 & 15.43$\pm$1.36 & 15.77$\pm$1.52 & 13.36$\pm$1.73 \\
SGPT & 23.67$\pm$2.62 & 23.29$\pm$2.64 & 23.30$\pm$2.78 & 22.28$\pm$2.78 \\
\bottomrule
\end{tabular}
\caption{Faithfulness analysis of explanation methods. AUPC (area under perturbation curve) on the STSb and BIOSSES semantic similarity datasets. Lower scores indicate better identification of features that decrease the Euclidean distance most effectively.}
\label{table:stsb_perturbation_scores_2}
\end{table*}

\section{Conservation} \label{app:conservation}
\begin{figure}[h]
    \centering\includegraphics[width=\linewidth]{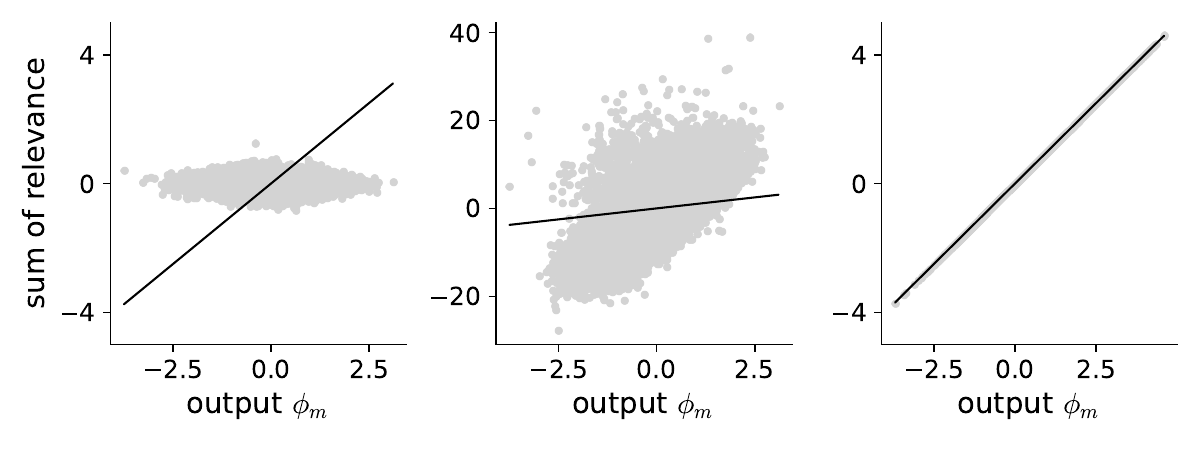}
    \caption{Conservation across 300 samples of the STSb test split on the SBERT model. Left: H$\times$P baseline. Center: BiLRP computation. Right: BiLRP computation with model biases set to zero.}
    \label{fig:conservation}
\end{figure}

Conservation of relevance is one important desired principle of post-hoc explanations. It ensures that relevance can not be created or disappear during the explanation process of the prediction. In Figure \ref{fig:conservation}, we show conservation of H$\times$P, BiLRP and BiLRP with biases set to zero for the SBERT model. We observe that for H$\times$P, the sum of predicted sentence embedding activations $\phi_m$ is not related to the sum of relevance scores at the input. For BiLRP, we observe a linear relationship of both quantities. The difference to full conservation (identity line) can be explained by layer biases that are not attributable by design.

\section{Multilingual experiments} \label{app:multilingual_triplets}
\begin{figure*}
    \scriptsize
    \centering
    \begin{multicols}{2}
    \raggedcolumns
    \includegraphics[width=\linewidth]{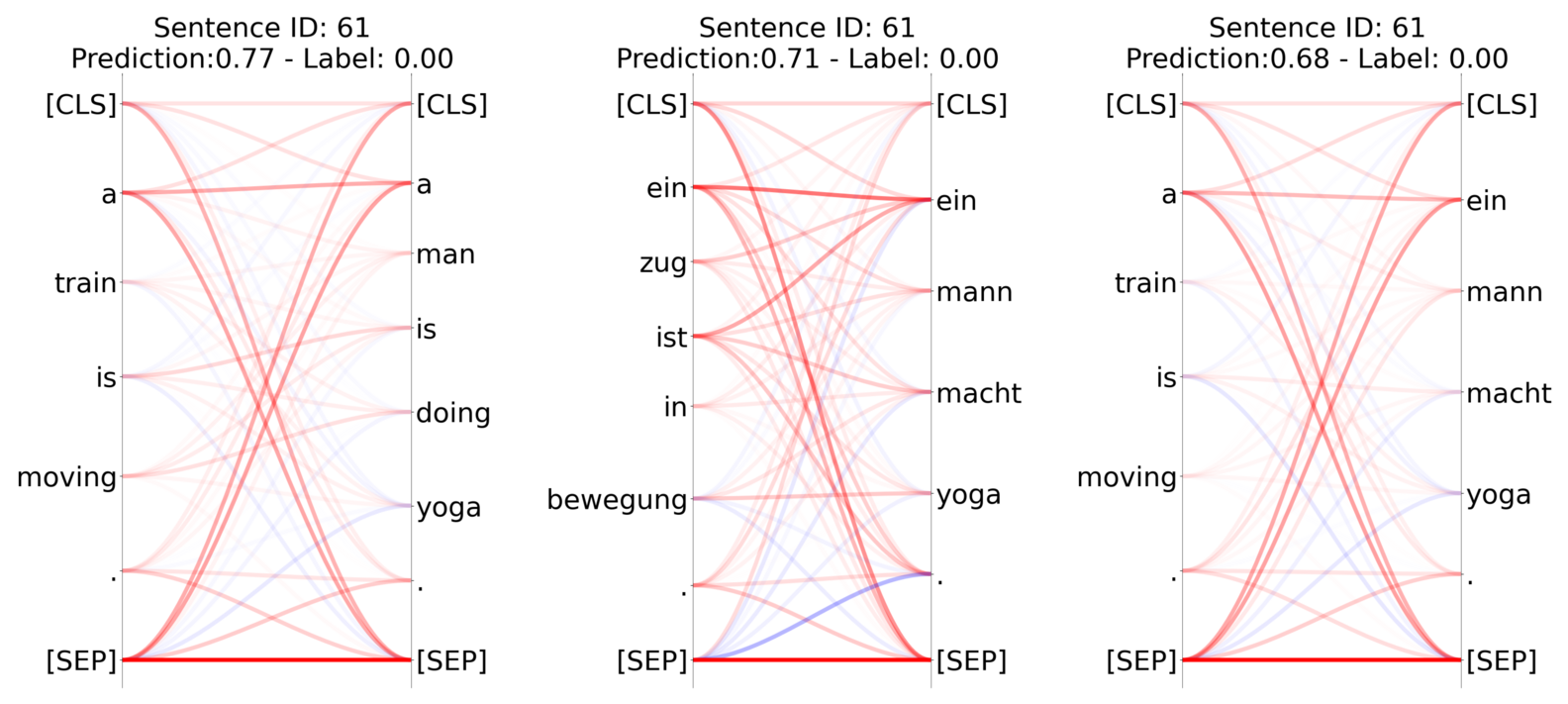}
    \includegraphics[width=\linewidth]{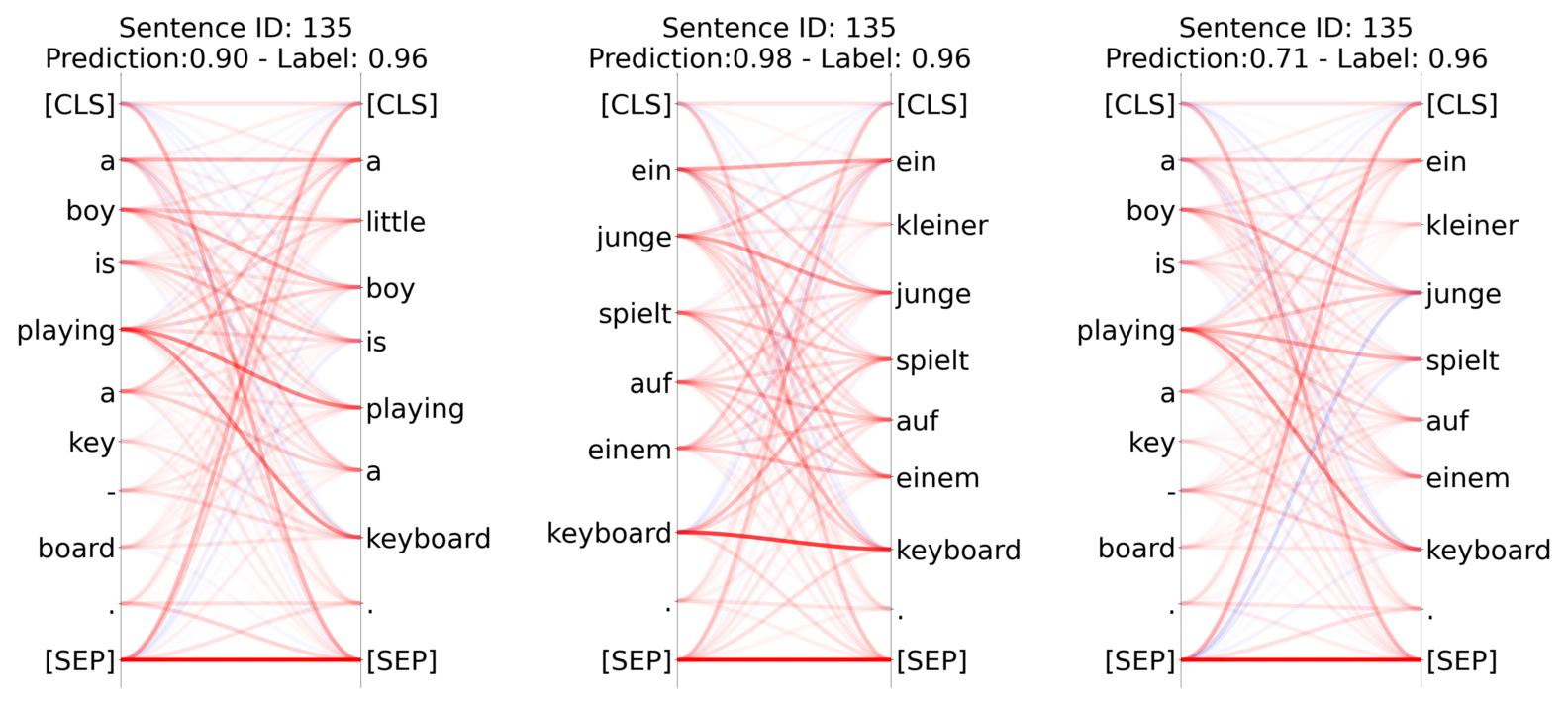}
    \includegraphics[width=\linewidth]{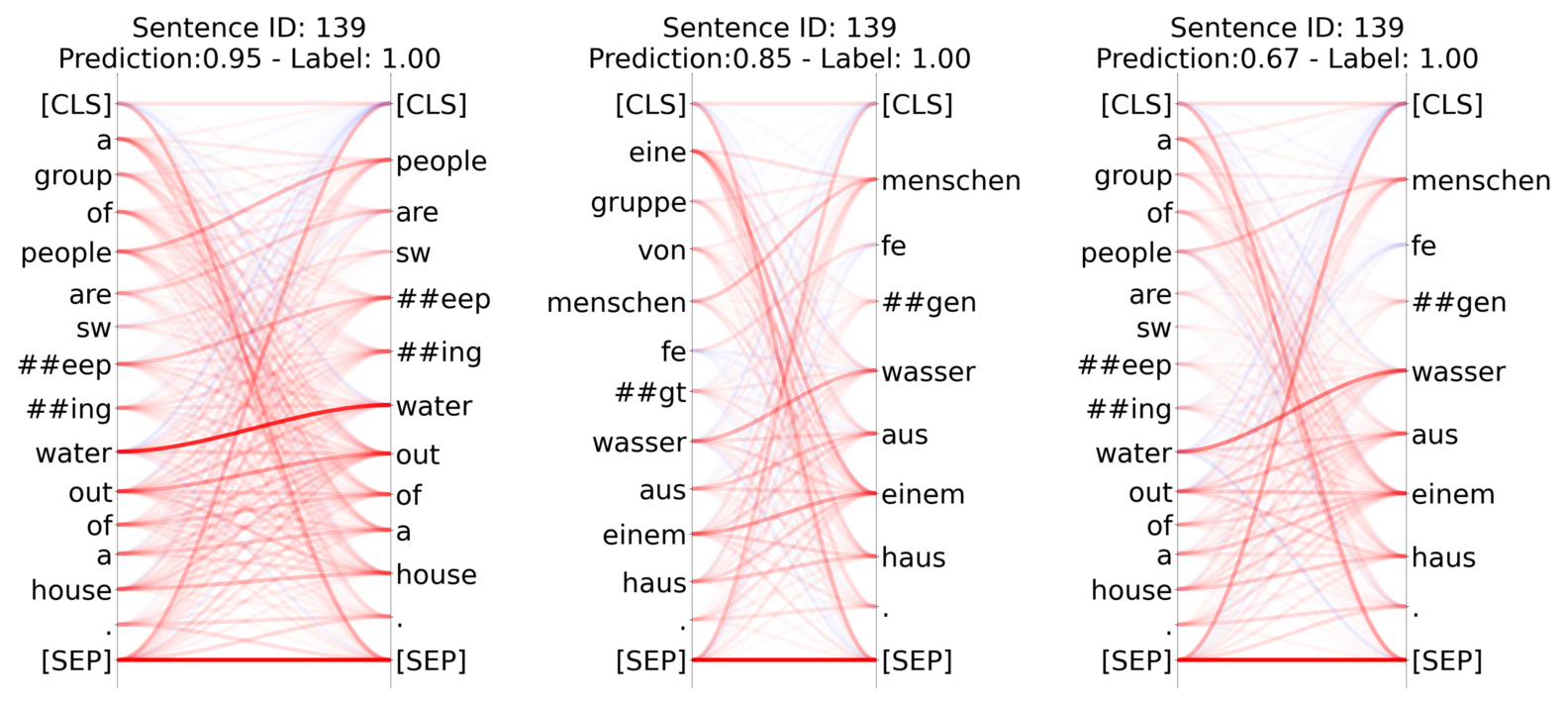}
    \includegraphics[width=\linewidth]{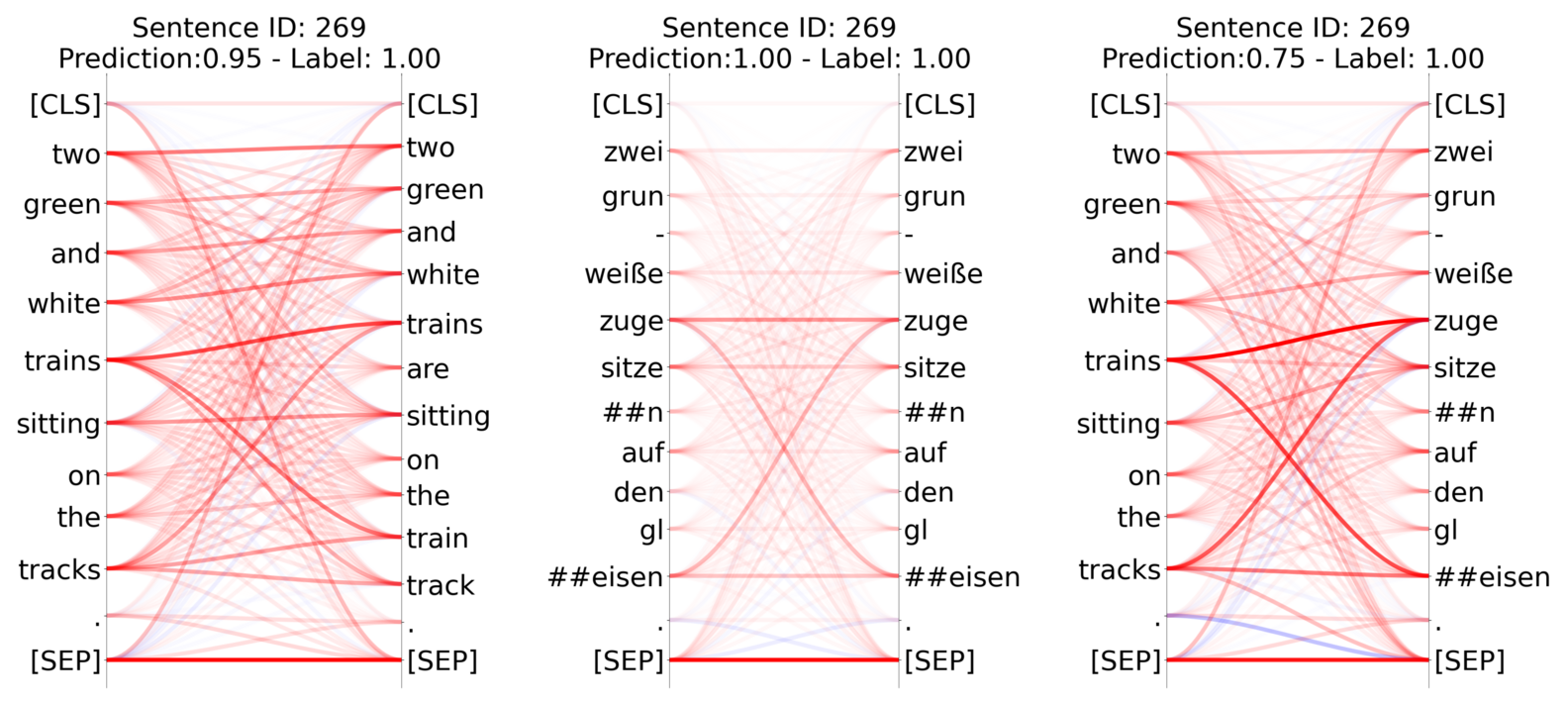}
    \includegraphics[width=\linewidth]{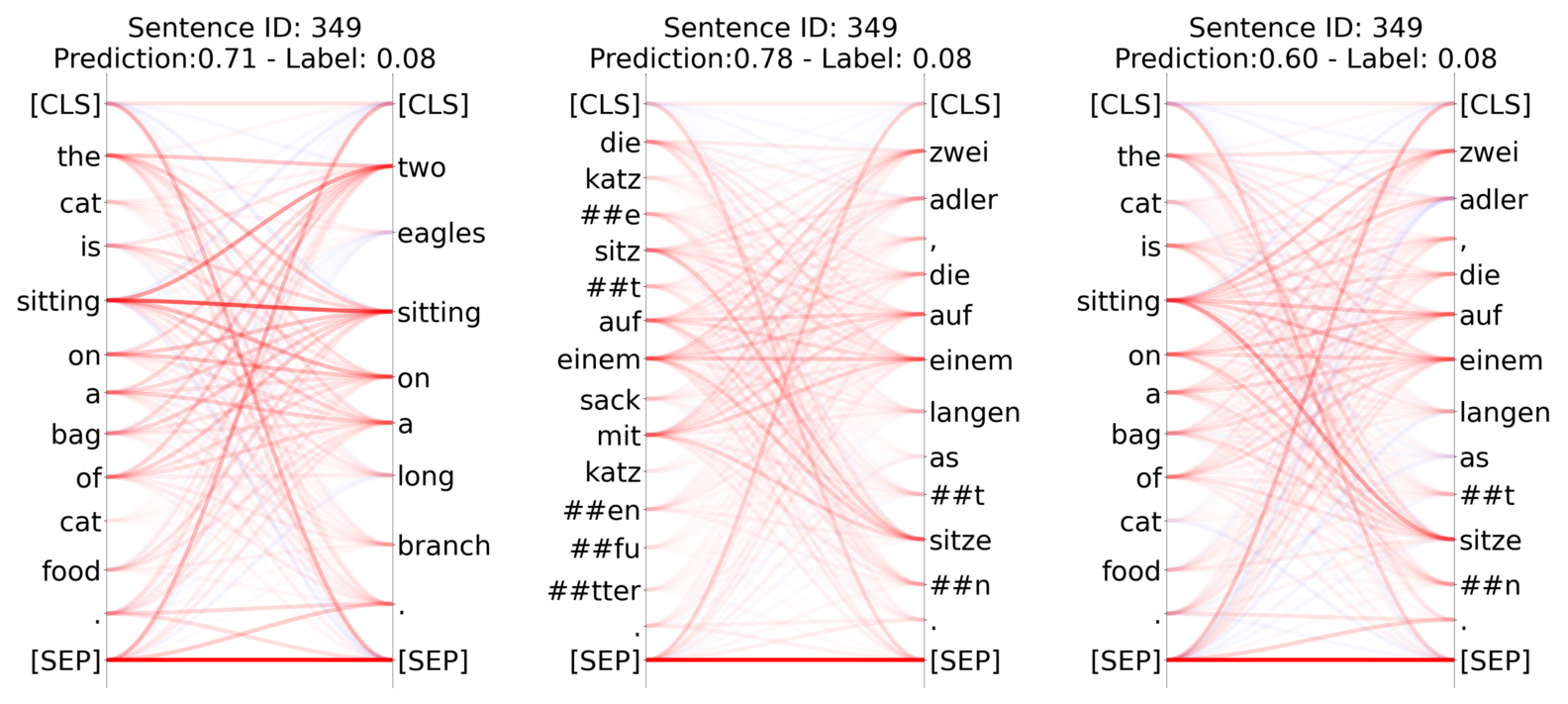}
    \includegraphics[width=\linewidth]{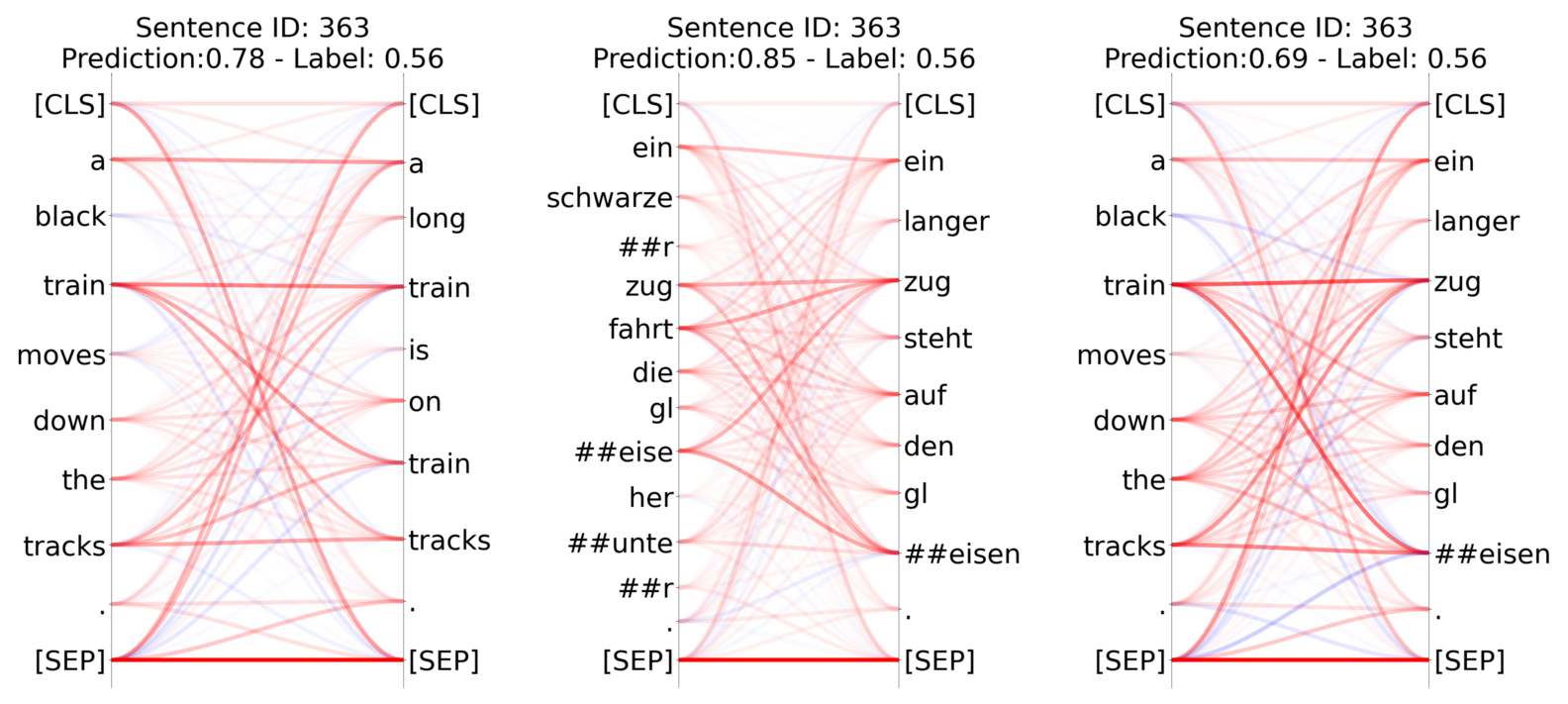}
    \includegraphics[width=\linewidth]{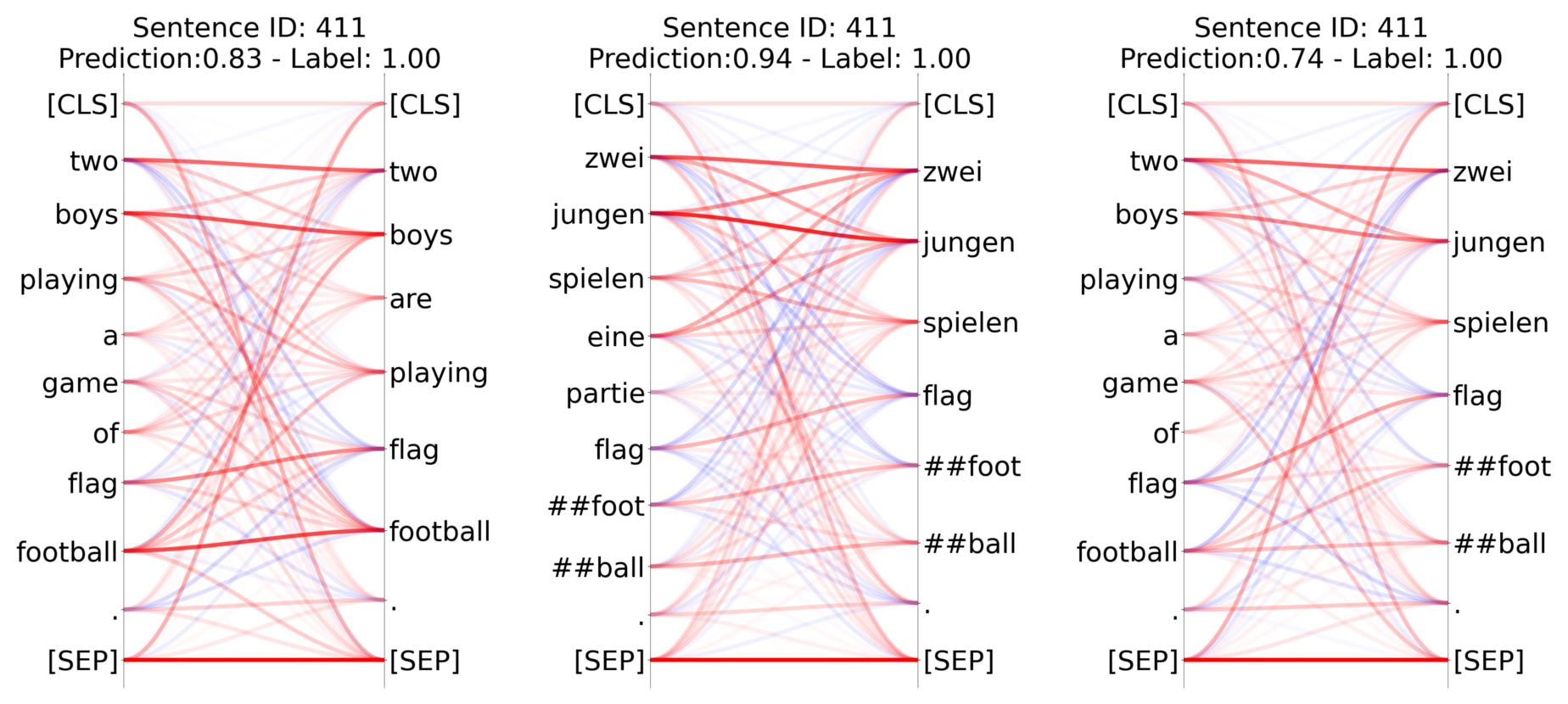}
    \includegraphics[width=\linewidth]{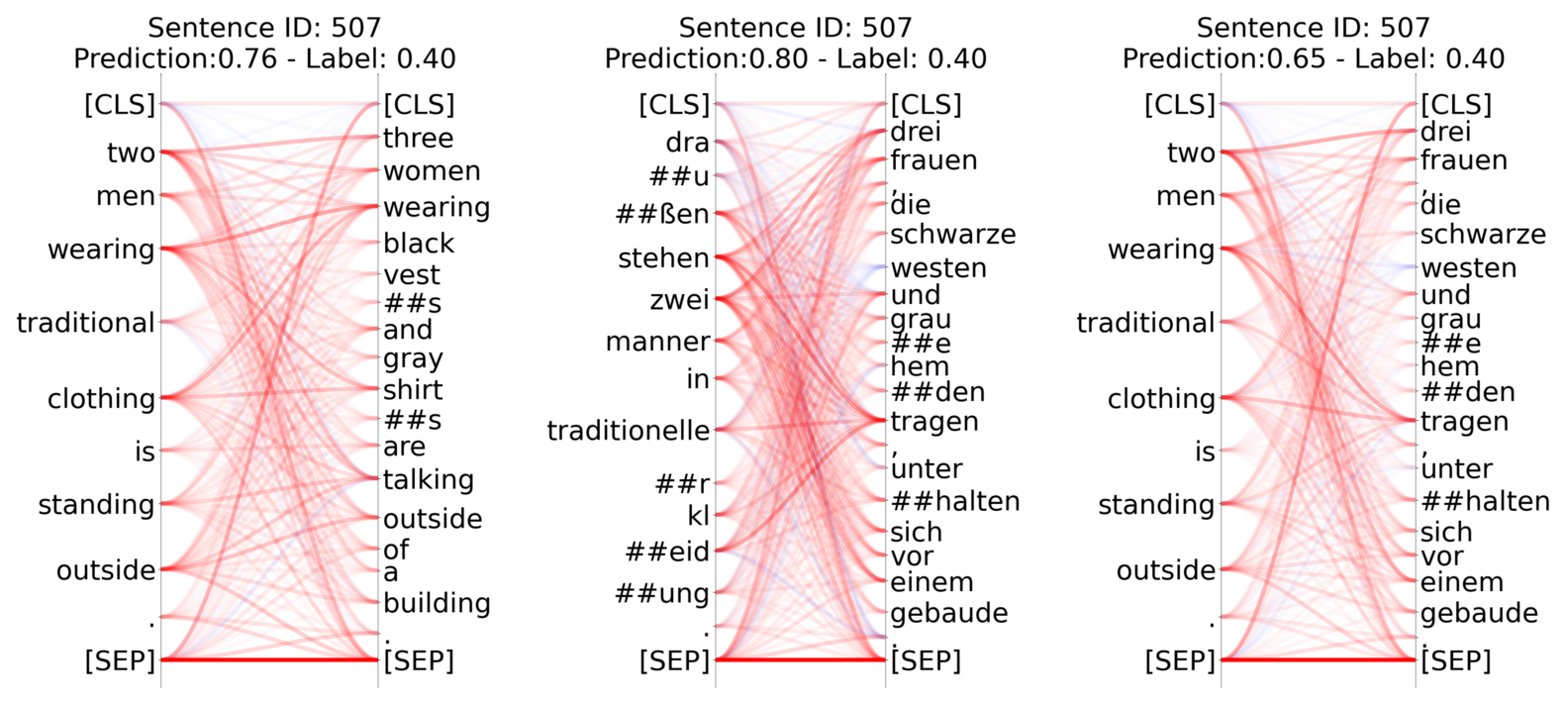}
    \end{multicols}
    \caption{Additional BiLRP explanations on mBERT for English, German and the mixed English-German samples. The samples are chosen as representatives of the model's matching strategies, depicting different similarity levels of the labels, similarity predictions, and cases where the two diverge the most.
    }
    \label{fig:app:multilingual_triplets}
\end{figure*}

In Figure \ref{fig:app:multilingual_triplets}, we provide additional examples of token interactions in the multilingual settings considered in Section \ref{sec:multilingual} of the main paper.

\begin{figure}[H]
    \centering\includegraphics[width=1.\linewidth]{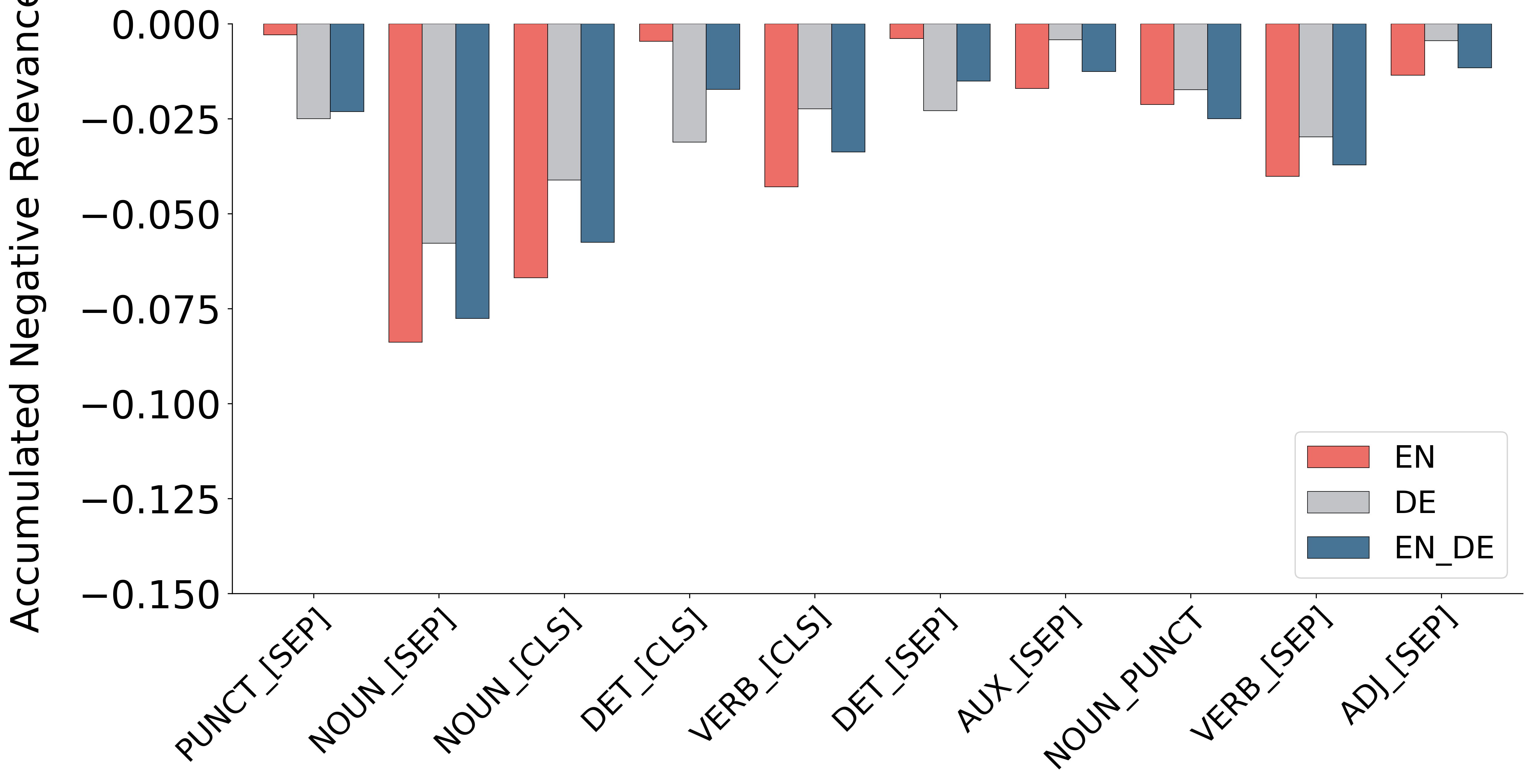}
    \caption{Comparison of relevant interaction between POS tags in a multilingual semantic similarity task. POS tags are selected based on largest difference of the negative accumulated relevance assigned to an interaction of POS tags between the mixed (EN-DE) setting and the monolingual (DE-DE, EN-EN) settings. }
    \label{app:fig:multilingual_neg}
\end{figure}

In analogy to the positively contributing POS interactions discussed in Section \ref{sec:multilingual}, we show POS interactions that contribute negatively in Figure \ref{app:fig:multilingual_neg}. Relevance scores are normalized by the maximum absolute value. Overall, we find that negative contributions are less strong with relevance magnitudes below 0.1 in comparison to the positive interactions that reach up to 0.8, as shown in Figure \ref{fig:multilingual_pos} in the main paper. 

\section{Additional POS Heatmaps}
In addition to the POS heatmaps in the main text, we provide the corpus-level analysis normalized by the number of POS-interaction occurrences. For this, we separate positive and negative relevance scores,  retrieve the mean relevance of each token pair (instead of the sum as in the main text), normalize the scores by dividing by the maximum absolute mean value. The resulting heatmaps present a complementary view to Figure \ref{fig:corpus_level} in the main paper, highlighting rare POS interactions. We observe that for \mbox{BERT + Mean Pooling}, the effect of the interactions between `[SEP]' tokens is more apparent, since it consistently ranks across the most relevant POS interaction. For example, BERT + CLS reacts strongly to  interactions between an interjection and a symbol (`nope-/'), while for SBERT we observe interactions between interjections to be of high relevance (`yes-yes' and `no-no'). Cases like this highlight the need for well-tuned models that base their similarity predictions on desired and plausible interactions.
\twocolumn

\begin{table*}
\begin{subfigure}{0.33\textwidth}
\caption*{\textmyfont{\textbf{a.}} BERT + CLS}
\includegraphics[width=\textwidth]{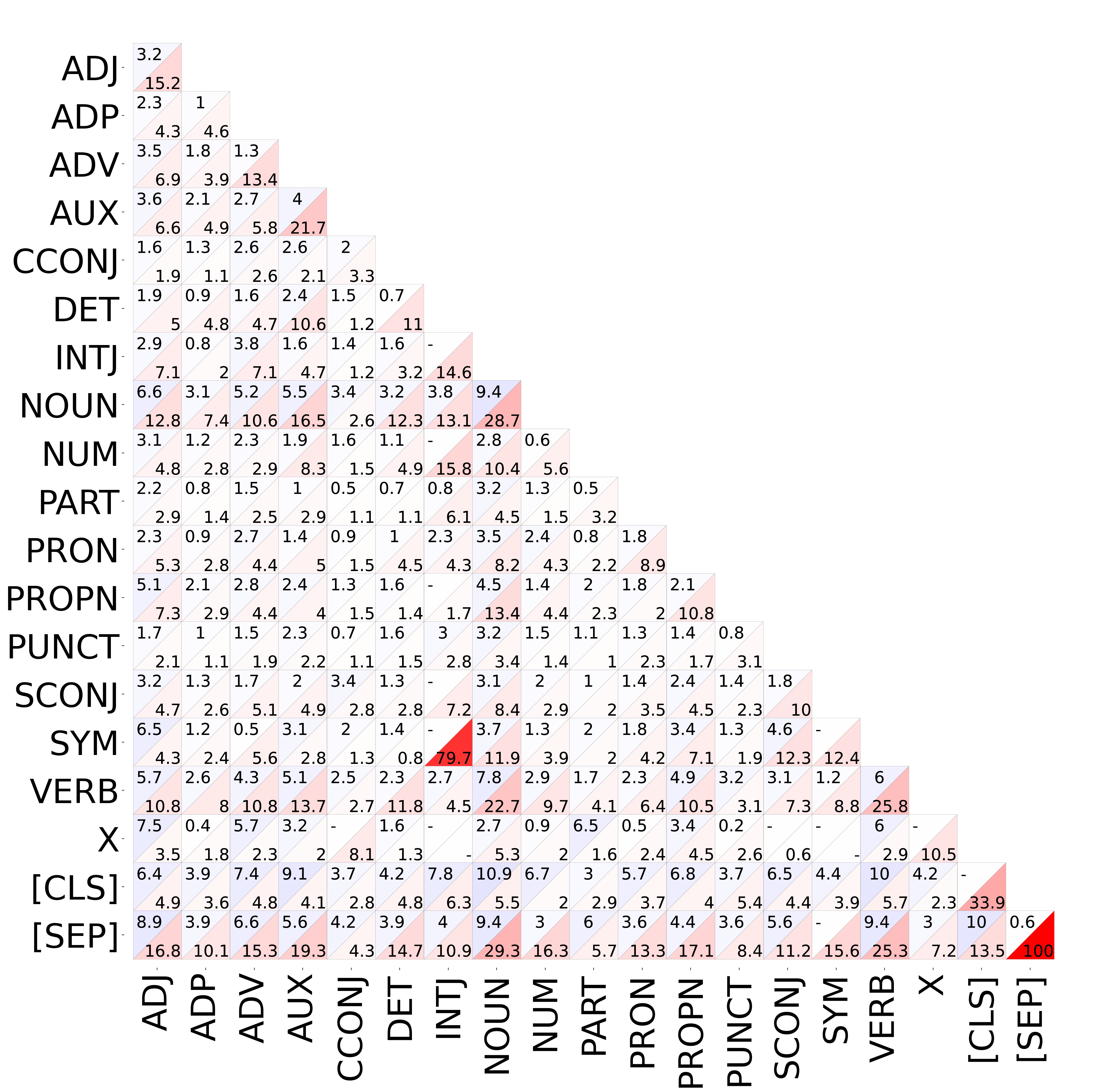}
\end{subfigure}\hfill
\begin{subfigure}{0.33\textwidth}
\caption*{\textmyfont{\textbf{b.}} BERT + Mean Pool.}
\includegraphics[width=\textwidth]{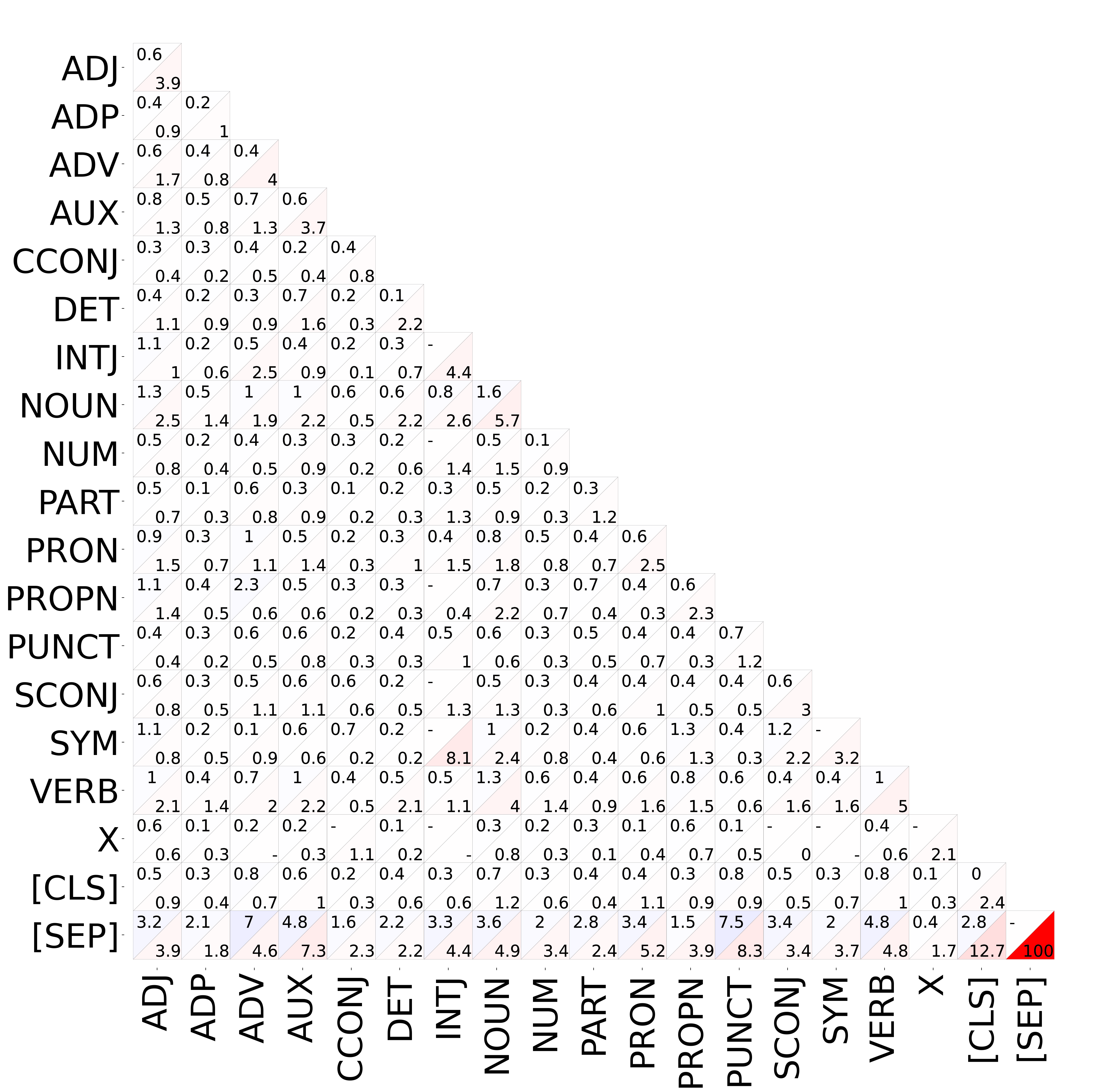}               
\end{subfigure}\hfill
\begin{subfigure}{0.33\textwidth}
\caption*{\textmyfont{\textbf{c.}} SBERT + Mean Pool.}
\includegraphics[width=\textwidth]{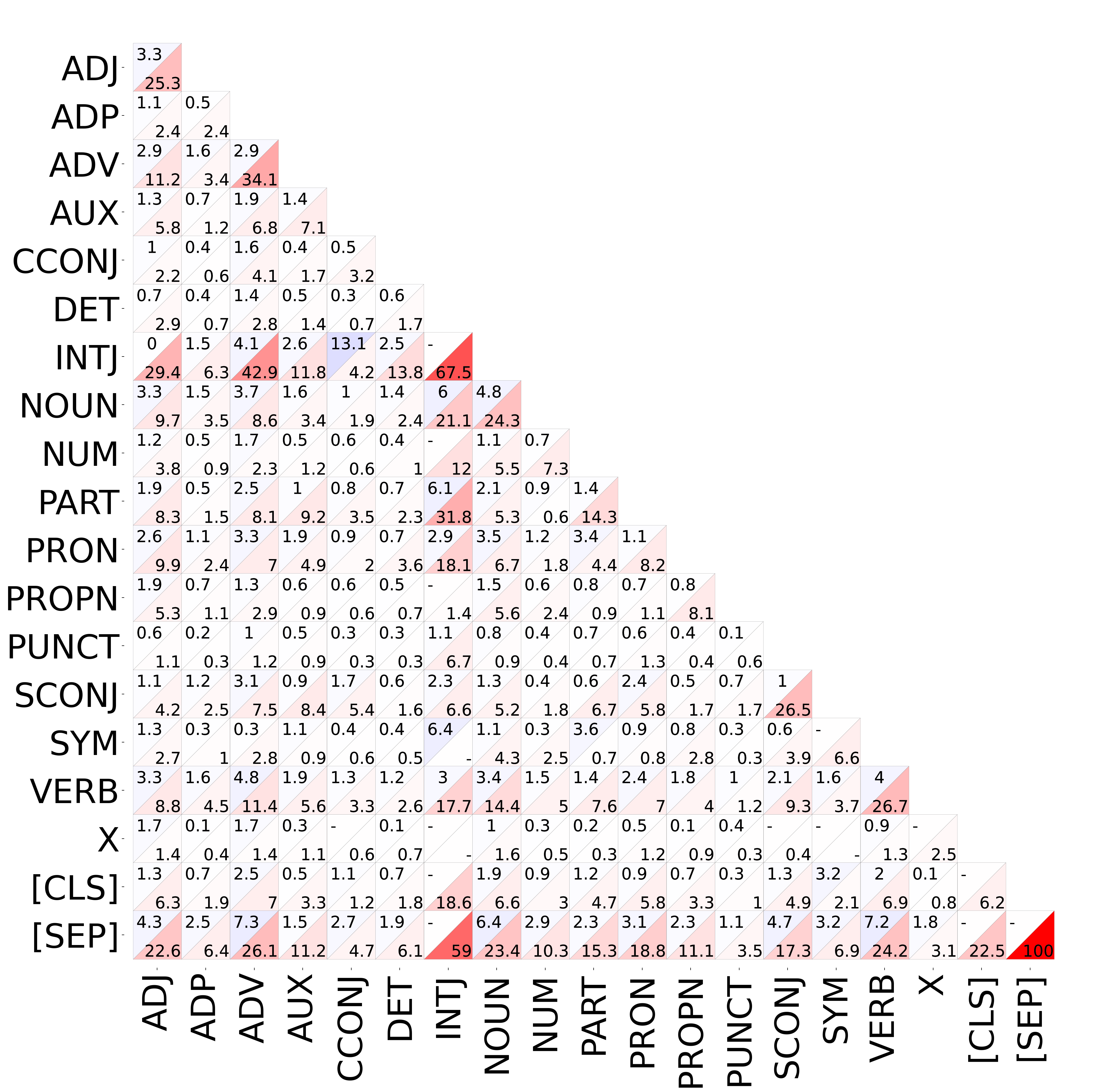}              
\end{subfigure}
\captionof{figure}{Alternative computation of the relevance scores, where the mean relevance of each token pair is retrieved instead of sum-pooling of relevance scores. Models range from (\textbf{a}) the least predictive  (BERT + CLS), to (\textbf{b}) moderately predictive (BERT + Mean Pooling), to (\textbf{c}) the most predictive (SBERT). 
}
\addtocounter{table}{-1}
\label{fig:app:corpus_level_alt}
\end{table*}

\end{document}